\documentclass[journal]{IEEEtran}
\usepackage{color}
\usepackage{multirow}
\usepackage{xspace}
\usepackage{bbding}
\usepackage[colorlinks,linkcolor=blue]{hyperref}
\newcommand*{\eg}{\textit{e.g}.\@\xspace}

\ifCLASSINFOpdf
\usepackage{graphicx}
\fi

\ifCLASSINFOpdf
\else
\fi

\begin{document}
%
\title{Learning Video Salient Object Detection Progressively from Unlabeled Videos}

\author{
	\IEEEauthorblockN{Binwei~Xu,
		Haoran~Liang\mbox{*},
		Wentian~Ni,
		Weihua Gong,
		Ronghua~Liang,~\IEEEmembership{Senior Member,~IEEE,}
		and~Peng~Chen}\\
    \IEEEauthorblockA{College of Computer Science and Technology,
    	Zhejiang University of Technology, Hangzhou, China}
\thanks{Corresponding author\mbox{*}: Haoran liang (email: haoran@zjut.edu.cn)}

}

%
%

%

\maketitle

\begin{abstract}
Recent deep learning-based video salient object detection (VSOD) has achieved some breakthrough, but these methods rely on expensive annotated videos with pixel-wise annotations, weak annotations, or part of the pixel-wise annotations. In this paper, based on the similarities and the differences between VSOD and image salient object detection (SOD), we propose a novel VSOD method via a progressive framework that locates and segments salient objects in sequence without utilizing any video annotation. To use the knowledge learned in the SOD dataset for VSOD efficiently, we introduce dynamic saliency to compensate for the lack of motion information of SOD during the locating process but retain the same fine segmenting process. Specifically, an algorithm for generating spatiotemporal location labels, which consists of generating high-saliency location labels and tracking salient objects in adjacent frames, is proposed. Based on these location labels, a two-stream locating network that introduces an optical flow branch for video salient object locating is presented. Although our method does not require labeled video at all, the experimental results on five public benchmarks of DAVIS, FBMS, ViSal, VOS, and DAVSOD demonstrate that our proposed method is competitive with fully supervised methods and outperforms the state-of-the-art weakly and unsupervised methods.
\end{abstract}


%
\IEEEpeerreviewmaketitle

\section{Introduction}\label{sec:introduction}
Video salient object detection (VSOD) is an important task toward intelligent video content understanding. It has wide applications in real-world vision tasks, such as video object segmentation~\cite{wang2015saliency}, video classification~\cite{rapantzikos2009spatiotemporal}, video summarization~\cite{evangelopoulos2009video}, video surveillance~\cite{hong2015online}, video compression~\cite{hadizadeh2013saliency}, and automatic driving~\cite{simon2009alerting}. Research~\cite{wang2015} also revealed the role of visual saliency in aiding in the diagnosis of autism. Therefore, VSOD has recently attached great attention to computer vision and multimedia processing communities. 
Unlike isolated images, video is more challenging with contextual associations and rich dynamic information. Existing VSOD methods mainly use the recurrent network~\cite{song2018pyramid,fan2019shifting,yan2019semi} or introduce optical flow information~\cite{li2019motion,ren2020tenet} to complement dynamic information. Although great progress has been made, the burden of collecting expensive video datasets with pixel-wise annotations still restricts the further development and application of VSOD. 

\begin{figure}[t]
	\centering
	\includegraphics[width=0.485\textwidth]{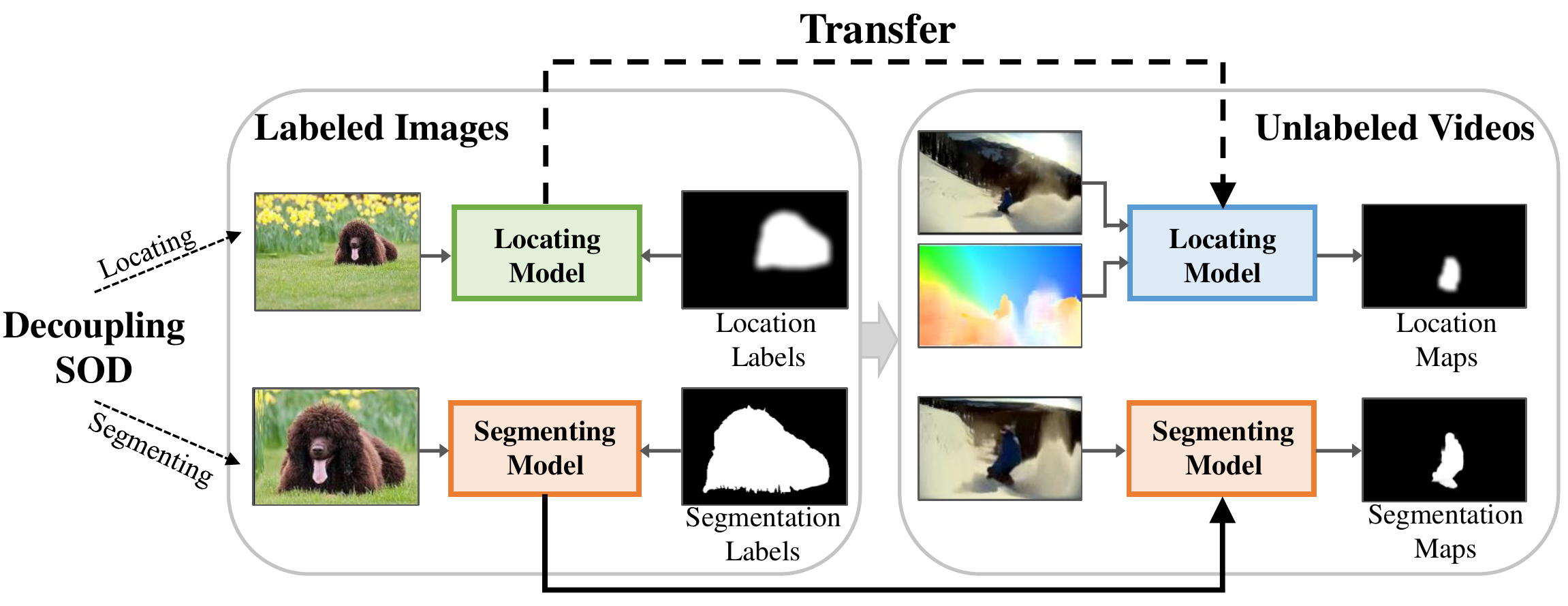} 
	\caption{
 Illustration of the overall flow of our proposed progressive VSOD method that does not use any video annotation. To transfer knowledge from images to videos efficiently, we decouple SOD into locating and segmenting. Specifically, we introduce optical flow information to compensate for the lack of motion information of VSOD during the locating process but retain the same segmenting process.
	}
	\label{fig:cover}
\end{figure}

Several recently fully supervised VSOD models~\cite{li2019motion,ren2020tenet} adopt a two-step training strategy. First, image salient object detection(SOD) dataset, \textit{e.g.}, DUTS~\cite{duts_wang} or MSRA10K~\cite{cheng2014global}, is employed to pretrain the model. After that, VSOD datasets, \textit{e.g.}, DAVSOD~\cite{fan2019shifting} and DAVIS~\cite{perazzi2016benchmark}, are used to fine-tune it. The main reason is that the image SOD dataset has rich scene diversity, whereas the VSOD dataset has limited scene diversity. In this way, the model has better generalization ability and can adapt to richer scenarios, producing better results. Concretely, although VSOD dataset, \textit{e.g.}, DAVSOD, has more frames for training than the image dataset, great redundancy is observed across the frames of each video. Hence, the limited valid scenes provided by the video limit the performance of the model. 
Training a robust model directly using a video dataset of such small size is difficult. The most intuitive, effective way to solve the problem of limited scene diversity is to collect more videos. However, time-consuming, expensive annotations for videos are the largest obstacle to the development of VSOD. 
Although unsupervised VSOD methods~\cite{rahtu2010segmenting,liu2014superpixel,xi2016salient,chen2017video} employ hand-crafted low-level features, they suffer from poor performance and work well only in several considered cases.
Recent research by~\cite{yan2019semi} proposed a semi-supervised saliency detection method by using sparsely labeled frames. It greatly reduces the manual labeling cost, but still requires 20\% pixel-wise annotations.
Zhao et al.~\cite{zhao2021weakly} designed a weakly supervised VSOD model via scribble to relieve the burden of pixel-wise labeling.
Although scribble annotations can effectively reduce time and capital cost compared with pixel-wise annotations, it still requires expensive, time-consuming labeling. The main reasons are not only the large size of video dataset, but also the reliance on high-cost fixation annotations, which are recorded by tracking the eye movements of many people using expensive eye trackers.
Although these learning-based methods can reduce the cost of video annotations, it’s difficult for them to deal with large-scale datasets that contain richer scenes in the coming future. In this paper, to overcome the challenges of limited scene diversity and high-cost labeling in the VSOD dataset, we aim to utilize scene-rich, effortless image SOD dataset for VSOD. At the same time, our method without using video annotations can perform better when collecting more videos with richer scenes.

\begin{figure}[t]
	\centering
	\includegraphics[width=0.48\textwidth]{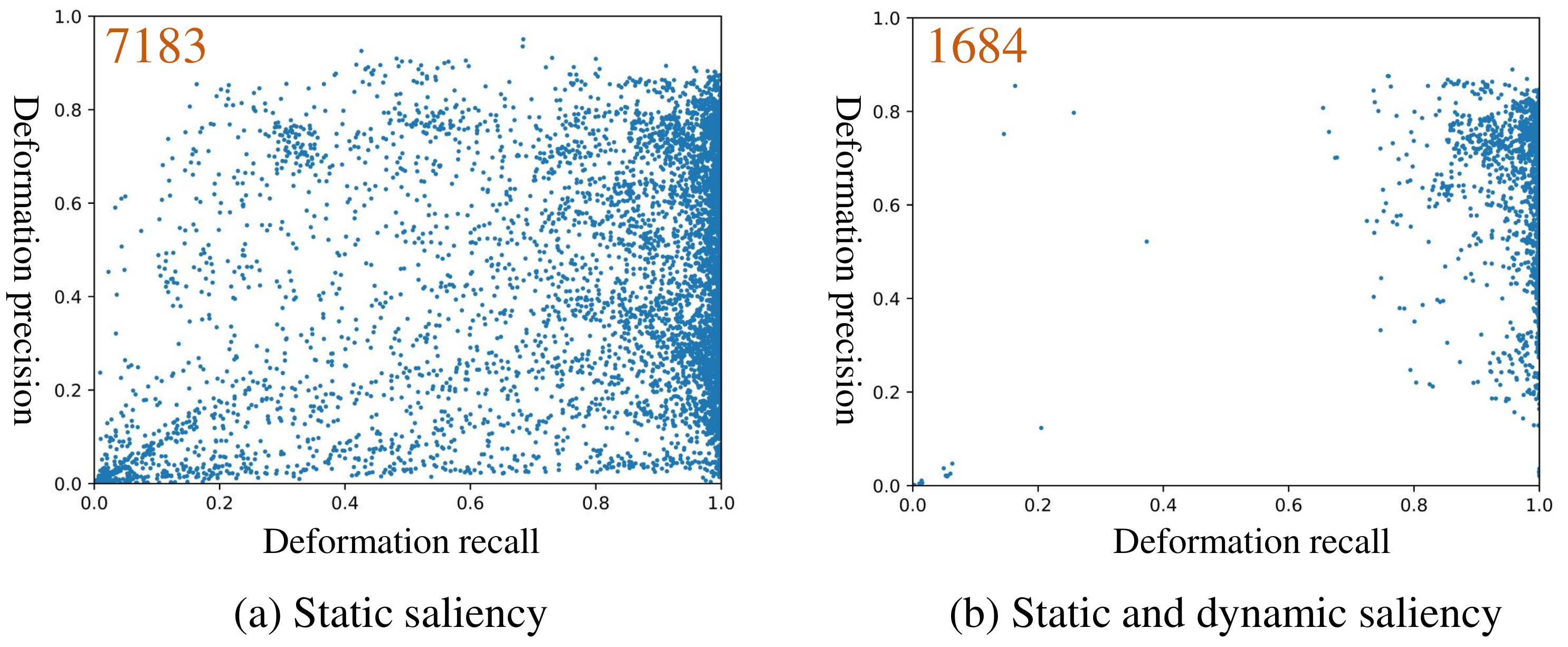} 
	\caption{Visualization of static saliency from DAVSOD training data with 7183 images by employing CLM~\cite{xu2021locate} trained on image datasets, and the 1684 results with static and dynamic saliency are selected. Deformation recall indicates the proportion of true salient objects in the predicted locating region and a higher value indicates that the prediction is more likely to contain complete salient objects. Deformation precision indicates the proportion of true salient objects in the predicted salient area and a quite low value indicates that no salient objects are included. Many points in (a) are distributed in the lower left corner, but most of the points in (b) are distributed in the rightmost side and the value of the distortion accuracy is more than 0.2.}
	\label{fig:precision_recall}
\end{figure}

Although image-based saliency detection models can be used to implement the task of VSOD directly, they perform poorly in distinguishing salient objects compared with video-based models.
In fact, salient object detection requires two tasks, locating and segmenting. Specifically, locating salient objects is to find the local region of salient objects from the global perspective of the whole image, whereas segmenting salient objects is to distinguish the boundary of salient objects from the local perspective of the regions that contain salient objects and their surrounding background. Although image SOD and VSOD tasks are different in locating salient objects, they share some degree of similarity because both require rigorous, fine segmentation of salient objects. 
Therefore, transferring knowledge to video saliency model by introducing models trained on image datasets is one of the feasible strategies.
Two key issues need to be solved, namely, decoupling salient object detection into locating and segmenting and locating video salient objects.
For the first problem, Xu et al.~\cite{xu2021locate} proposed a novel progressive architecture for image salient object detection that consists of coarse locating module (CLM) and fine segmenting module (FSM)\footnote{CLM and FSM mentioned in this paper are based on our previous work published on AAAI conference. Code and data are publicly available at our \href{https://github.com/bradleybin/Locate-Globally-Segment-locally-A-Progressive-Architecture-With-Knowledge-Review-Network-for-SOD}{Project page}.}.
These two modules effectively decouple the functions of locating and segmenting salient objects, which provides a good solution.
Compared with locating salient objects in images, detecting salient objects in videos is different. Temporal information can be a critical cue for identifying the salient objects in videos. Salient objects of a video can be detected by image-based SOD models. However, this straightforward idea leads to inaccurate location of video salient objects because of ignoring object motion (\eg, temporal information) in the consecutive frames. Previous works~\cite{ren2020tenet,li2019motion} added an optical flow branch to leverage motion feature and achieve promising results, which indicates that the introduction of optical flow images can effectively capture temporal information and solve the above problem.
The above observations inspire us to think whether regions with appearance saliency (static saliency) and motion variation (dynamic saliency) can be considered locations of salient objects in a video frame.
Fig.~\ref{fig:precision_recall} verifies our idea, that is, locating the salient objects in the video only by single static saliency is insufficient, we must take into account both static and dynamic saliency. (detailed in Section.~\ref{sec::generate-high}).

Based on the above problems and thoughts, in this work, we propose a novel VSOD method via progressive framework by using image SOD data and unlabeled videos. 
As shown in Fig.~\ref{fig:cover}, after decoupling SOD into locating and segmenting independently, we design a progressive framework to transfer knowledge efficiently from images to videos based on the differences in locating and similarities in segmentation between SOD and VSOD.
Specifically, we employ the CLM trained on the image SOD dataset that can predict the static location map and the dynamic location map from frames and their optical flow images, respectively. 
Then, we design a spatiotemporal location label generation method by measuring the similarity of the static location map and the dynamic location map.
Next, we propose an optical flow tracking method to obtain the spatiotemporal location maps of the same salient objects in adjacent frames, which are also considered as the video salient object location maps.
Furthermore, we propose a two-stream locating network to locate video salient objects by using those frames and corresponding spatiotemporal location maps.
Based on the similarity between image SOD and VSOD on the segmentation of salient objects, we directly apply the fine segmentation model trained on the image SOD dataset to achieve accurate segmentation. 
Through such a two-stage approach of coarse locating and fine segmenting, knowledge learned from the image dataset is efficiently transferred  to the VSOD task.

To demonstrate the effectiveness of our proposed method, we implement experiments on five popular benchmarks named DAVIS~\cite{perazzi2016benchmark}, FBMS~\cite{ochs2013segmentation}, ViSal~\cite{wang2015consistent}, VOS~\cite{li2017benchmark}, and DAVIS~\cite{perazzi2016benchmark}, and visualize some examples of saliency maps. Further, we conduct ablation studies to verify the reliability of important modules.
Our main contributions are as follows:
\begin{enumerate}
	\item For the first time, we design a progressive deep framework for VSOD based on image SOD data and unlabeled videos. Our method does not require video annotations at all, but the results are competitive with those of existing fully supervised VSOD methods and outperform those of the state-of-the-art weakly or unsupervised methods.
	\item We propose a novel algorithm for generating spatiotemporal location labels that consists of generating high-saliency location labels and tracking salient objects in adjacent frames, which can transfer knowledge learned from images indirectly for locating video salient objects.

	\item We propose a two-stream locating network that bridges static and dynamic information to predict the salient object location maps for videos, which can adapt to the data structure of sparse spatiotemporal location labels and add important motion information through optical flow images.

\end{enumerate}

\begin{figure*}[t]
	\centering
	\includegraphics[width=0.9\textwidth]{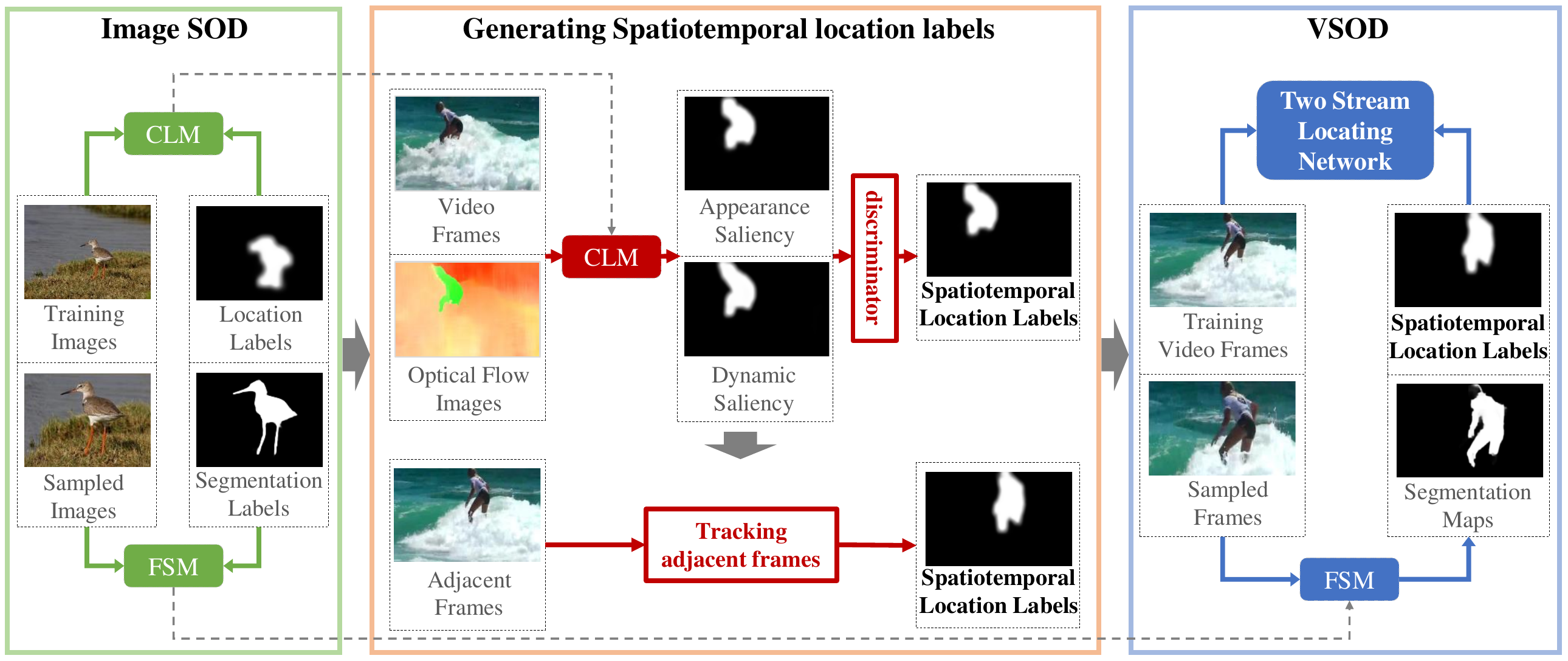} 
	\caption{Overview of our proposed method. It is composed of image SOD, generating spatiotemporal location labels, and VSOD.
	The image SOD consists of a CLM used to generate spatiotemporal location labels and a FSM employed for the final video salient object segmentation. Spatiotemporal location labels are generated by applying trained CLM. Based on the generated spatiotemporal location labels, a two-stream locating network that can locates salient regions of all frames is trained. Finally, VSOD is accomplished by the fine-trained two-stream locating network and FSM.}
	\label{fig:framework}
\end{figure*}

\section{Related Work}
\subsection{Image Salient Object Detection}

Early methods detect salient objects by employing hand-crafted priors with low-level visual cues such as color~\cite{borji2012exploiting}, texture~\cite{yan2013hierarchical}, and contrast~\cite{perazzi2012saliency}. Clearly, hand-crafted features are insufficient to capture high-level semantics, and many works~\cite{zhao2015saliency,luo2017non} use a convolutional neural network (CNN) to learn deep features for finding salient objects. Later, several works~\cite{liu2018picanet,zhao2019pyramid} found that the shallow layers are even more important in capturing salient feature, and they further integrated additional features from multiple layers to leverage global and local contexts for improving model performance. 
Among them, Hou et al.~\cite{hou2017deeply} introduced short connections and combined features from different layers to generate saliency maps. Zhang et al.~\cite{zhang2017amulet} derived a resolution-based feature combination module and a boundary-preserving refinement strategy. Hu et al.~\cite{hu2018recurrently} aggregated deep features to exploit the complementary saliency information between the multi-level features and the features at each individual layer. Zhang et al.~\cite{zhang2018bi} formulated a bi-directional message passing model to select features for integration. Zhang et al.~\cite{zhang2018progressive} developed an attention-guided network to select and integrate multi-level information. Li et al.~\cite{li2018contour} presented a contrast-oriented deep neural network, which adopts two network streams for dense and sparse saliency inference. Chen et al.~\cite{chen2018reverse} leveraged residual learning and reverse attention to refine saliency maps. Zhang et al.~\cite{zhang2019salient} designed a symmetrical CNN to learn the complementary saliency information and presented a weighted structural loss to enhance the boundaries of salient objects. Xu et al.~\cite{xu2021locate} proposed a novel progressive architecture for image salient object detection that consists of coarse locating module and fine segmenting module.
Although these methods make great progress, they are inapplicable to VSOD without considering temp-spatial information and contrast information.

\subsection{Full Supervised Video Salient Object Detection}
The mainstream fully supervised video salient detection models mainly focus on how to effectively explore the temporal and spatial information of annotated data.
Wang et al.~\cite{wang2017video} employed FCN to take adjacent pairs of frames as input.
Chen et al.~\cite{chen2018scom} presented a spatiotemporal constrained optimization model, which exploits spatial and temporal cues, as well as a local constraint, to achieve a global saliency optimization.
To capture a wider range of temporal information, Song et al.~\cite{song2018pyramid} designed a recurrent network framework for the fast detection of video salient object.
Li et al.~\cite{li2018flow} proposed enhancing the temporal coherence at the feature level by exploiting motion information and sequential feature evolution encoding.
Several methods also explore human attention mechanisms to select salient regions.
Fan et al.~\cite{fan2019shifting} proposed a saliency-shift-aware module to learn human attention shift.
Gu et al.~\cite{gu2020pyramid} designed a constrained self-attention model to capture temporal information.
As moving objects are usually salient in videos, Li et al.~\cite{li2019motion} developed a multi-task motion guided network to learn SOD in still images and motion saliency detection in optical flow images.
Chen et al.~\cite{chen2021exploring} presented a new temp-spatial network, in which the key innovation is the design of a temporal unit.
Zhang et al.~\cite{zhang2021dynamic} proposed a dynamic context-sensitive filtering network to adapt to dynamic variations and introduce spatial information over time.
Despite the great progress in VSOD, time-consuming, expensive annotations remain the largest obstacle to the further development of VSOD.

\subsection{Weakly/Semi/Un-supervised Video Salient Object Detection}
Several unsupervised VSOD methods employ hand-crafted low-level features.
Rahtu et al.~\cite{rahtu2010segmenting} advocated a salient object segmentation method based on combining a statistical framework with a conditional random field model.
Based on the super pixel representation of video frames, Liu et al.~\cite{liu2014superpixel} extracted motion histograms and color histograms at the super pixel level and the frame level, and integrated them into the spatiotemporal saliency map.
Xi et al.~\cite{xi2016salient} investigated how to identify the background priors for a video and detected the salient objects by employing the background priors.
Chen et al.~\cite{chen2017video} proposed temp-spatial saliency fusion and low-rank coherency guided saliency diffusion for video saliency detection.
Wang et al.~\cite{wang2015saliency} used spatial boundaries and temporal motion edges as indicators of foreground object locations.
However, these traditional methods suffer from poor performance in real-world application and work well only in several considered cases.
To alleviate the issue of relying on expensive, laborious pixel-wise dense annotations, few weakly supervised and semi-supervised learning-based methods started to emerge.
Tang et al.~\cite{tang2018weakly} presented a spatiotemporal cascade neural network architecture, in which two fully CNNs are cascaded to evaluate the visual saliency from spatial and temporal cues.
Li et al.~\cite{li2020plug} proposed to retrain pretrained saliency detection models weakly during testing by using pseudo labels.
Yan et al.~\cite{yan2019semi} presented a semi-supervised saliency detection method by using sparsely labeled frames. It greatly reduces the manual labeling cost, but still requires 20\% pixel-wise annotations, which would be a large project for large video datasets with much richer scenes.
Zhao et al.~\cite{zhao2021weakly} designed a weakly supervised VSOD model via scribble to relieve the burden of pixel-wise labeling. Although scribble annotations can effectively reduce time and capital cost compared with pixel-wise annotations, it still requires expensive, time-consuming labeling.
Different from previous learning-based approaches, which rely on part of the fully annotations or weak annotations, our method is based on effortless, scene-rich image dataset without using any annotated video label.


\section{Methodology}
\label{sec::method}
In this section, we elaborate the details of our proposed method. Fig.~\ref{fig:framework} illustrates the framework of our method. It consists of three main steps: image salient object detection, generating spatiotemporal location labels, and VSOD. First, CLM and FSM are trained by image SOD dataset DUTS-TR~\cite{duts_wang}. CLM aims to highlight areas that contain salient objects. After locating salient objects, FSM achieves their accurate segmentation. Second, fine-trained CLM is employed to identify static salient areas from video frames and dynamic salient regions from their optical flow images. When static salient regions and dynamic salient regions include the same objects in a frame, these objects are regarded as video salient objects. Based on the similarity between static salient regions and dynamic salient regions, the high-saliency frames can be identified, and their video salient areas can be located. According to the continuity of salient objects in adjacent frames, salient regions that contain the same salient objects in adjacent frames can be located by the adjacent frame tracking method. Finally, a two-stream locating network is proposed to learn video salient object location by using high-saliency frames and their adjacent frames with their optical flow images as inputs and salient regions predicted by CLM as spatiotemporal location labels. 
After locating salient objects, the fine segmenting results can be generated by fine-trained FSM.

\begin{figure}[t]
	\centering
	\includegraphics[width=0.485\textwidth]{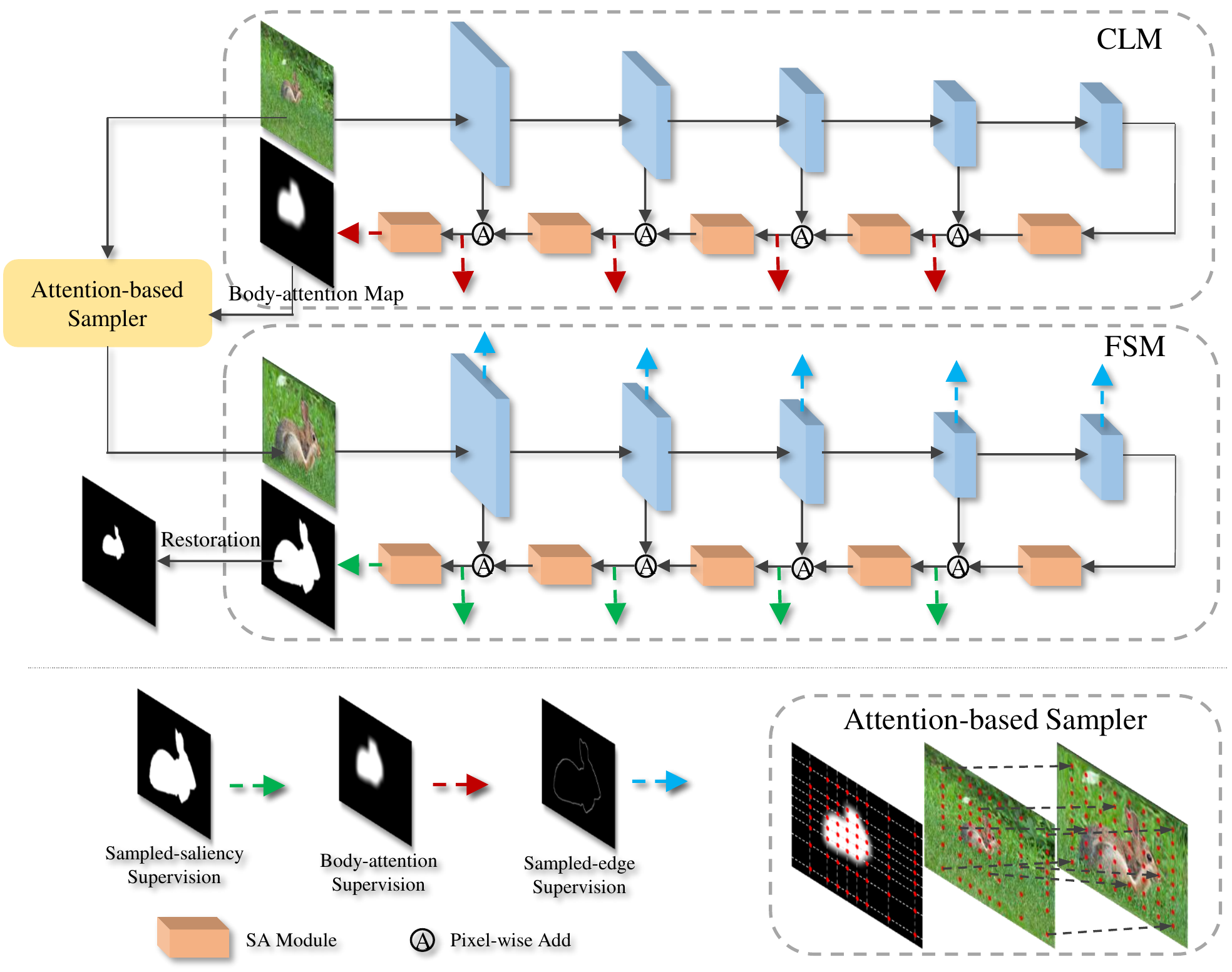} 
	\caption{Illustration of the flow of image salient object detection.}
	\label{fig:image_detection}
\end{figure}
\subsection{Image Salient Object Deteciton}
The task of VSOD includes not only distinguishing the region where the salient objects are but also accurately segmenting the salient objects. Compared with the image salient objects, the video salient objects change due to the introduction of motion information. However, the accurate segmentation of the objects is similar to that in the image.
Thus, we split the image saliency detection into two parts, CLM and FSM, which can help VSOD take full advantage of the knowledge of location and segmentation learned in the image, respectively.
This step consists of three parts: CLM, FSM, and attention-based sampler.
As shown in Fig.~\ref{fig:image_detection}, CLM is a network that locates salient objects, which is based on feature pyramid networks (FPNs)~\cite{fpn_lin}. 
The body-attention location map is treated as the label of CLM to guide the model to focus on rough regions that contain salient objects, which are generated by the dilation operation and Gaussian blur operation. 
To accomplish the task of precise segmentation, body-attention location maps output by CLM are employed to increase the resolution of the regions related to salient objects in an image, which can magnify the details of salient objects. Such a processed image is treated as the input for FSM, which guides the FSM to focus on the segmentation of salient objects. 
The structure of the FSM is the same as the CLM, except that intermediate edge supervisions are applied in the decoder process to guide the network to retain more details.
The attention-based sampler~\cite{zheng2019looking} is proposed in the fine-grained classification task to magnify the details of salient objects. 
The main idea is that regions of attention-based maps with high attention value are sampled more intensively by calculating the mapping function between the coordinates of the original image and the sampled image.
The restoration process reinstates the predicted sampled segmentation result to its original form by this calculated mapping function.

\subsection{Generating Spatiotemporal Location Labels}
\subsubsection{Generating High-saliency Location Labels}
\label{sec::generate-high}



Unlike static images, videos have rich motion characteristics. 
Specifically, salient objects in a static image are mainly determined by their static appearance (including texture, color, and semantics), whereas object saliency in the video is influenced by static appearance and the motion.
For example, if a video has no evident moving object, static spatial information will play a decisive role. However, in static videos, such as surveillance video, moving objects are more attractive to human visual attention. Generally, in most videos, static appearance information and motion information play indispensable roles.
Therefore, accurately establishing the relationship between static information and dynamic information can improve the performance of VSOD.

Several special objects in the video have appearance saliency and dynamic saliency. That is, 
these objects can attract human visual attention in a static frame and keep distinct movement in consecutive frames.
These objects are denoted as spatiotemporal salient objects. When every object in a video frame is a spatiotemporal salient object, this frame is treated as a high-saliency frame.
As shown in Fig.~\ref{fig:framework}, spatiotemporal location label generating consists of appearance saliency prediction branch, dynamic saliency prediction branch, and high-saliency frame discriminator. For the static image branch, we employ CLM trained on the image dataset to find static salient regions. In the dynamic branch, we estimate the optical flow of video frames by~\cite{flow_gao} and render it according to~\cite{render_butler}. In this way, motion information is visualized on a single image. When the object has motion distinct from the background, its area in the rendered optical flow image has distinguishable colors and clear boundaries. As a result, we can easily detect dynamic salient regions by CLM.
After acquiring the predicted body-attention location maps of static frames and optical flow images, the high-saliency frame discriminator is used to determine whether their salient regions are similar. We design a strict discriminant condition to ensure the high quality of generated high-saliency frames. Intersection over union (IOU) is to measure the similarity of the generated body-attention maps of static frames and optical flow images: 
\begin{equation}
IOU\;=\;\frac{\sum_{r=1}^H\sum_{c=1}^WS(r, c)G(r, c)}{\sum_{r=1}^H\sum_{c=1}^W\lbrack S(r, c)+G(r,c)-S(r, c)G(r, c)\rbrack}
\end{equation}
where $G(r, c)$ is the optical flow saliency of the pixel $(r,c)$ and $S(r, c)$ is the predicted probability of static saliency. $H$ and $W$ are the height and the weight of the video frame, respectively.
When the IOU value of a frame higher than the threshold $T$, it is considered a high-saliency frame.
For these high-saliency frames, we use the static image salient areas predicted by CLM as video salient object location maps. Because the dynamic salient areas may not contain the complete object due to the local motion of the salient object, whereas the static image salient areas are obtained by the trained CLM that has learned the integrity of the object from the image SOD dataset, which has a high probability of containing complete objects. Ultimately, we employ these frames and their corresponding salient object location maps as a part of training data for subsequent learning of two-stream locating networks.


To verify whether regions with static saliency and dynamic saliency can be considered locations of salient objects, the results of static saliency from DAVSOD training data with 7183 images are visualized by CLM and 1684 results with static and dynamic saliency are selected, as shown in Fig.~\ref{fig:precision_recall}.
CLM uses labels after the inflation and Gaussian blurring operations as training data; thus, the results predicted by CLM are blurred but can fully contain salient objects.
Each blue point represents the result of one image.
The horizontal coordinate is deformation recall (D-Recall), similar to recall, which indicates the proportion of true salient objects in the predicted locating region, and a higher value indicates that the prediction is more likely to contain complete salient objects. 
The vertical coordinate is deformation precision (D-Precision), similar to precision, which indicates the proportion of true salient objects in the predicted salient area, and a quite low value indicates that no salient objects is included. D-Recall and D-Precision can be computed as follows:
\begin{equation}
D-Recall\;=\;\frac{\sum_{r=1}^H\sum_{c=1}^WL(r, c)T(r, c)}{\sum_{r=1}^H\sum_{c=1}^W T(r, c)}
\end{equation}
\begin{equation}
D-Precision\;=\;\frac{\sum_{r=1}^H\sum_{c=1}^WL(r, c)T(r, c)}{\sum_{r=1}^H\sum_{c=1}^W L(r, c)}
\end{equation}
where $L(r, c)$ is the prediction of the pixel $(r,c)$ by CLM, and $T(r, c)$ is the ground truth. $H$ and $W$ are the height and the weight of the video frame, respectively.
Many points in Fig.~\ref{fig:precision_recall}a are distributed in the lower left corner, but most of the points in Fig.~\ref{fig:precision_recall}b are distributed in the rightmost side and the value of the distortion accuracy is more than 0.2. It clearly proves that locating the salient objects in the video by single static saliency is insufficient, but the salient objects can be located by dynamic saliency and static saliency.

\begin{figure}[t]
	\centering
	\includegraphics[width=0.485\textwidth]{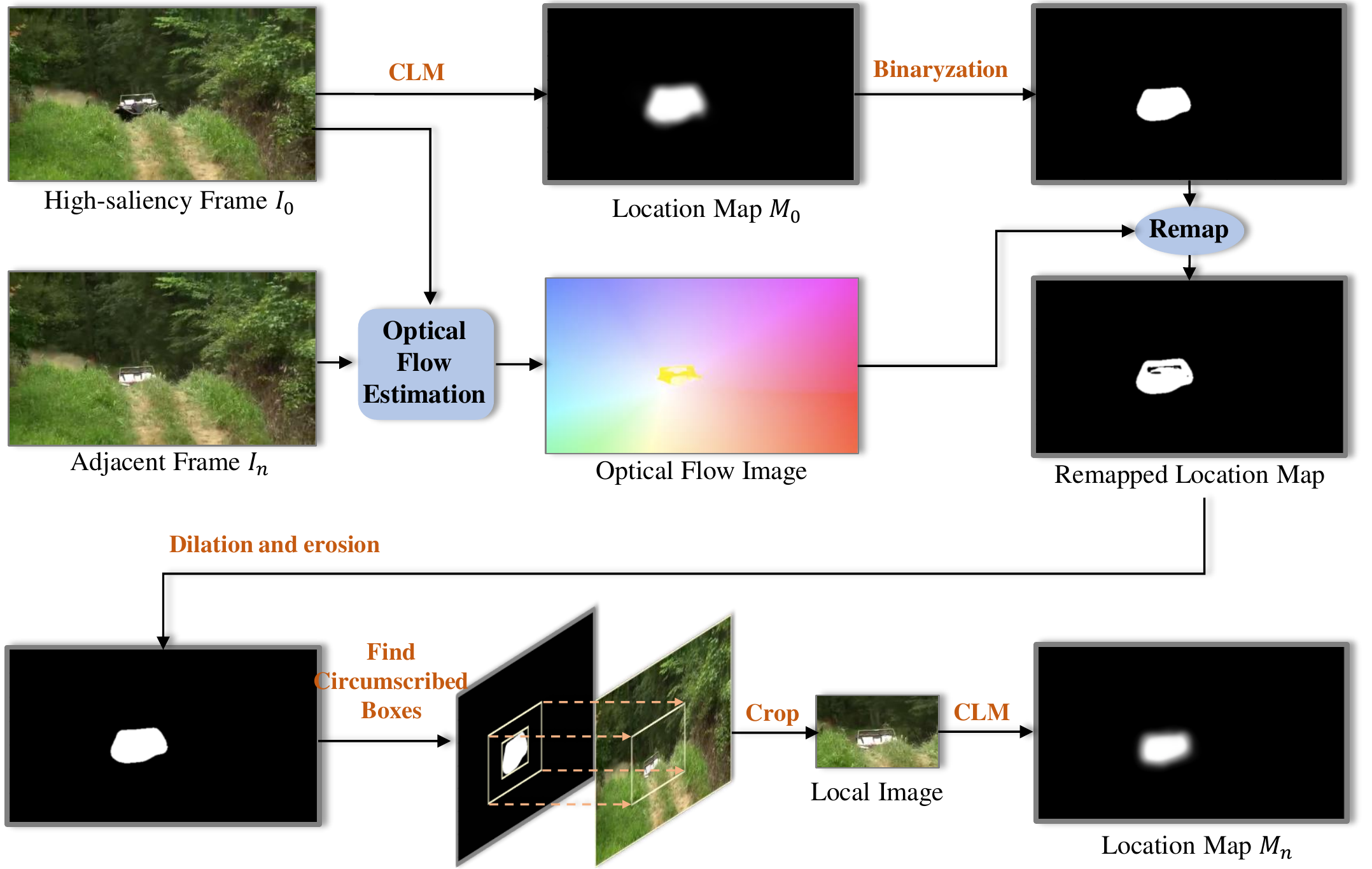} 
	\caption{Illustration of tracking salient objects in adjacent frames.}
	\label{fig:neighber}
\end{figure}
\subsubsection{Tracking Adjacent Frames}
After obtaining high-saliency frames with corresponding salient object location maps, salient object location maps of the adjacent frames are generated based on the corresponding optical flow information.
If the training data of the two-stream locating network consists of only high-saliency frames with corresponding location maps, the trained model performs well only on such kind of frames but poorly on other frames without high-saliency objects. 
Depending on the continuity of the video, the video salient objects in the adjacent frames of a high-saliency frame are highly probable to be the same as those in the high-saliency frame.
Therefore, we can avoid above-mentioned issues by adding these frames to the training data. 

As shown in Fig.~\ref{fig:neighber}, given a high-saliency frame $I_0$ and its corresponding spatiotemporal   location map $M_0$, the adjacent frames tracking method is used to calculate spatiotemporal location maps $M_n$ of adjacent frames $I_n$, where n indicates the first six frames and the last six frames of $I_0$. 
Based on~\cite{teed2020raft}, we can obtain the optical flow information between $I_0$ and adjacent frame $I_n$. 
The map records the displacement of objects. 
Based on the spatiotemporal location map of the high-saliency frame and the optical flow image, we calculate salient object location maps of the adjacent frames by using the remap function of OpenCV~\cite{2008Learning}. 
However, it may result in a remapped location map that does not contain the complete object and such location labels lack the incorrect shape of salient objects that may degrade the performance of subsequent training.
To avoid the above problems, we crop out wider regions that contain remapped salient objects and reuse CLM to calculate its salient object location map.
Specifically, We first dilate and erode the remapped location map, and then find the circumscribed boxes of the connected domains and double the size of the circumscribed boxes to include complete objects.
Based on the coordinates of these boxes, the local images containing the salient objects are generated by cropping the adjacent frame.
Finally, we employ CLM to predict the location maps of these local images and map them to the original size.
In addition, a situation where a adjacent frame can be a high-saliency frame or have different spatiotemporal location maps generated by different high-saliency frames can occur. For this situation, we select the spatiotemporal location map based on the proximity principle. 

\begin{figure*}[t]
	\centering
	\includegraphics[width=1\textwidth]{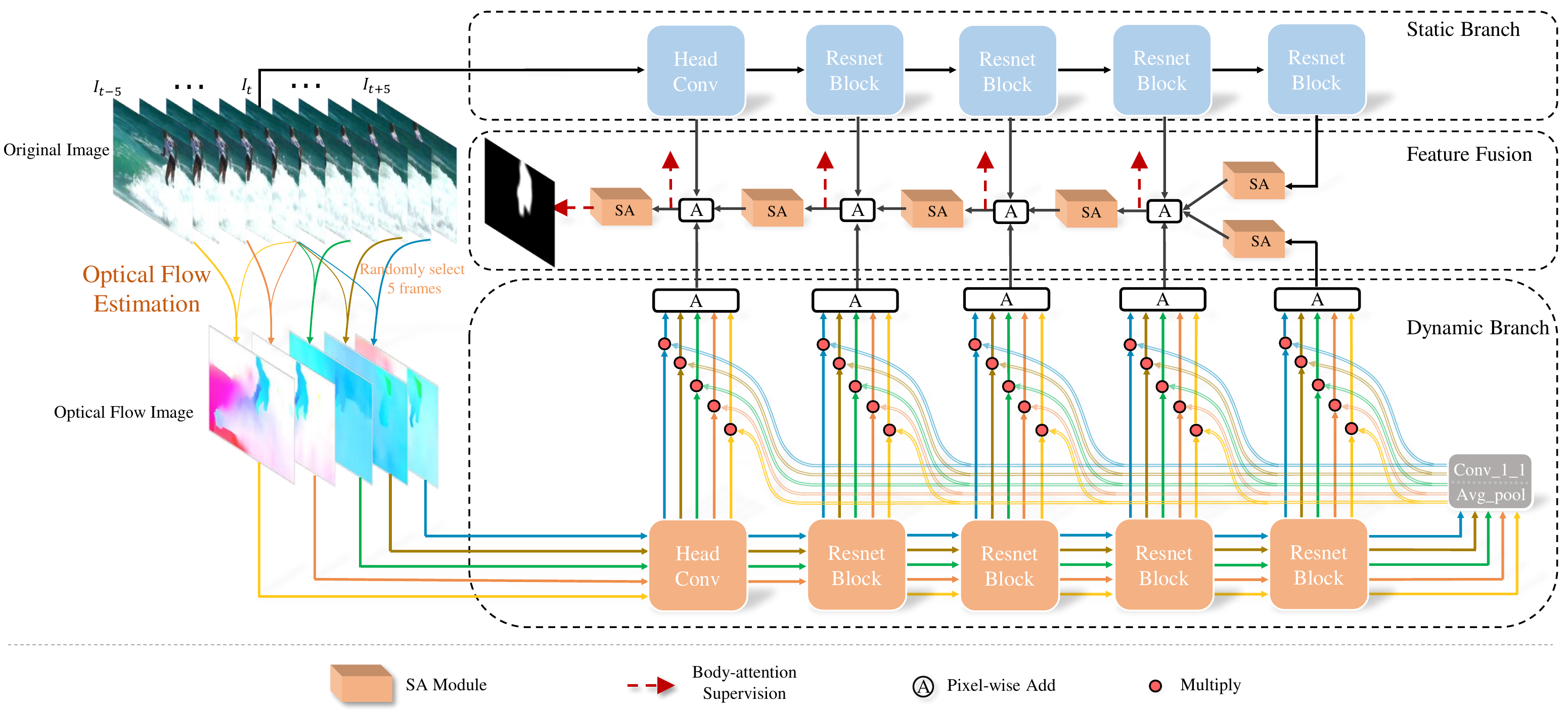} 
	\caption{Detailed illustration of our two-stream locating network, which consists of static feature extraction branch, dynamic feature extraction branch, and feature fusion module.}
	\label{fig:network}
\end{figure*}

\subsection{Two-stream locating Network}
\label{sec::two-stream}

After obtaining high-saliency frames and their neighboring frames as the video salient object location dataset, we propose a two-stream locating network to learn the location of salient objects. This network can adapt to the data structure of our sparse spatiotemporal location labels effectively and introduce important motion information through the optical flow images.
As shown in Fig.~\ref{fig:network}, it is mainly composed of static feature extraction branch, dynamic feature extraction branch, and feature fusion module.  The static branch and the dynamic branch are based on CLM~\cite{xu2021locate} pretrained on the SOD dataset as the backbone, which can provide multi-scale features with rich semantic information. 
The input of the static branch is the RGB image and the inputs of the dynamic branch are five optical flow images. To obtain these optical flow images, we randomly choose five images from the first five frames and the last five frames to calculate optical flow images with the current frame. If the optical flow branch only inputs one optical flow frame, the result of the network is easily affected by the quality of the optical flow image. However, the five optical flow images contain more or less different motion information, which can provide richer information. In addition, our random selection is similar to data augmentation, which can improve the generalization ability of the model.

The feature fusion module not only needs to combine static features and dynamic features to obtain more comprehensive semantic information, but also needs to recover the spatial size of feature maps and predict high-resolution saliency maps by fusing high-level and low-level features. 
The basic structure of this module is the same as that of the decoder of FPN~\cite{fpn_lin} that gradually integrates features layer by layer through the top-down transmission to obtain comprehensive information. The main difference is that Side-out Aggregation (SA) Module~\cite{poolnet_liu} is introduced after feature fusion of each layer. 
In the feature fusion module, we use output feature maps of the static branch and the dynamic branch as inputs, denoted as ${C_{s1}, C_{s2}, C_{s3}, C_{s4}, C_{s5}}$ and ${C_{m1}, C_{m2}, C_{m3}, C_{m4}, C_{m5}}$ respectively. 
A weight is added to each feature map because of the difference in the quality of each input optical flow image in the dynamic branch. Specifically, the feature maps of the last output layer are compressed to one dimension by a 1x1 convolution, and the weight of a certain optical flow image is obtained by average pool operation. 
These weights are then multiplied with the feature maps of layers corresponding to the optical flow image. Finally, the feature maps of the five optical flow images are added as the input to the feature fusion module.
In addition, we add intermediate supervisions after each fusion to guide the network to retain only the helpful information related to salient objects.
Feature integration consists of four steps. In the first fusion, $C_{s4}$, $C_{m4}$, the output of $C_{s5}$ that passes through the SA module, and the output of $C_{m5}$ that passes through the SA module are combined by the pixel-wise sum and a $3\times3$ convolution. In the next series of fusions,  corresponding static features, corresponding motion features, and features from the previous layer that passes through the SA module are combined by pixel-wise sum and a $3\times3$ convolution. In addition, intermediate supervisions are added after each fusion to guide the network to retain only the helpful information related to salient objects.

\subsection{Loss Function}
Similar to the CLM, two-stream locating network aims to predict the distribution of salient objects rather than pixel-wise value. Sen~\cite{47jia2020eml} proposed a combined loss function that consists of linear correlation coefficient (CC), Kullback-Leibler divergence (KLD), and normalized scanpath saliency (NSS). We modify NSS to meet our goal. 
The loss function of body-attention supervision is formulated as follows:
\begin{equation}
l_{b}=NSS'+CC'+KLD
\end{equation}
where \begin{math}NSS'\end{math} and \begin{math}CC'\end{math} indicate the variant of NSS and CC, respectively. We denote predicted maps as $P$ and body-attention maps as $Q$.
Additionally, we pick up pixels with values of 255 in $Q$ to generate a new ground truth as $F$, which is the area with a high probability of salient objects.

NSS is employed to measure the average normalized values of \textit{P} at the eye fixation points in fixation prediction \textit{F}~\cite{48peters2005components}, which highlights the importance of these pixels. Our salient object location focuses on is the pixels with high value in $F$. \begin{math}NSS'\end{math} is shown as follows:
\begin{equation}
NSS'(P,F)=\frac1N\sum_i(\frac{F-\mu(F)}{\sigma(F)}-\frac{P-\mu(P)}{\sigma(P)})\times F_i
\end{equation}
where \textit{i} is the \begin{math}i^{th}\end{math} pixel, $N$ indicates the number of pixels with high value in $F$, and \begin{math}\sigma(\cdot)\end{math} and \begin{math}\mu(\cdot)\end{math} are the standard deviation and the mean of the input, respectively.
The CC metric is commonly used to measure the linear correlation between two variables. 
\begin{equation}
CC(P,Q)=\frac{\sigma(P,Q)}{\sigma(P)\times\sigma(Q)}
\end{equation}
where \begin{math}\sigma(P,Q)\end{math} is the covariance of \textit{P} and \textit{Q}.
\begin{math}CC'\end{math} is formulated as follows:
\begin{equation}
CC'(P,Q)=1-\frac{\sigma(P,Q)}{\sigma(P)\times\sigma(Q)}.
\end{equation}
KLD is used to measure the similarity between two distributions and formulated as follows:
\begin{equation}
KLD(P,Q)=\sum_iQ_i\log(\frac{Q_i}{P_i+\varepsilon})
\end{equation}
where \begin{math}\varepsilon\end{math} is a regularization term.

The total loss of two-stream locating network ($L_{tln}$) is formulated as follows: 
\begin{equation}
L_{tln}=\sum_{i=1}^5l_{b}^i
\end{equation}
where $l_{b}^i$ indicates the loss of the $i^{th}$ intermediate body-attention supervision. Fig.~\ref{fig:network} shows all supervisions.

%

\begin{table*}[]
	\scriptsize
	\begin{center}
		\setlength{\tabcolsep}{1mm}{
			\begin{tabular}{cc|ccccccccccccc|ccccc}
				\hline
                &                          & \multicolumn{13}{c}{Fully Sup. Models}                                                                                                                                                                                                            & \multicolumn{5}{|c}{Weakly/unsup. Models}                     \\
				&                          & PoolNet & EGNet & PAKRN                        & SCOM  & MBNM  & PDB   & FGRN  & MGA                          & RCRN                         & SSAV                         & PCSA  & STVS                         & DCFNet                       & SSOD  & GF    & SAG   & WS    & Ours                         \\
				\multirow{-3}{*}{Dataset} & \multirow{-3}{*}{Metric} & \cite{poolnet_liu}  & \cite{zhao2019egnet}& \cite{xu2021locate} & \cite{chen2018scom}& \cite{li2018unsupervised}& \cite{song2018pyramid}& \cite{li2018flow} & \cite{li2019motion}& \cite{yan2019semi}& \cite{fan2019shifting}& \cite{gu2020pyramid}& \cite{chen2021exploring}
			    & \cite{zhang2021dynamic} & \cite{zhang2020weakly} & \cite{wang2015consistent}& \cite{wang2015saliency} & \cite{zhao2021weakly}&                         \\
				\hline
				\hline
				& $S_{measure}$            & 0.702   & 0.719 & 0.732                        & 0.599 & 0.637 & 0.698 & 0.693 & 0.741                        & 0.741  & {\textcolor{red} {0.755}} & 0.741 & 0.744                        & 0.741                        & 0.672    & 0.553    & 0.565   & 0.694    & {\textcolor{red} {0.744}} \\
				& $F_{max}$                & 0.592   & 0.604 & 0.632                        & 0.464 & 0.520 & 0.572 & 0.573 & 0.643                        & 0.654  & 0.659                        & 0.656 & 0.650                        & {\textcolor{red} {0.660}} & 0.556    & 0.334    & 0.370   & 0.593    & {\textcolor{red} {0.659}} \\
				\multirow{-3}{*}{DAVSOD} & MAE                      & 0.089   & 0.101 & 0.089                        & 0.220 & 0.159 & 0.116 & 0.098 & 0.083                        & 0.087  & 0.084                        & 0.086 & 0.086                        & {\textcolor{red} {0.074}} & 0.101    & 0.167    & 0.184   & 0.115    & {\textcolor{red} {0.085}} \\
				\hline
				& $S_{measure}$            & 0.902   & 0.946 & 0.952                        & 0.762 & 0.898 & 0.907 & 0.861 & 0.940                        & 0.922  & 0.942                        & 0.946 & {\textcolor{red} {0.954}} & 0.952                        & 0.853    & 0.757    & 0.749   & 0.883    & {\textcolor{red} {0.954}} \\
				& $F_{max}$                & 0.891   & 0.941 & {\textcolor{red} {0.956}} & 0.831 & 0.883 & 0.888 & 0.848 & 0.936                        & 0.907  & 0.938                        & 0.941 & 0.953                        & 0.953                        & 0.831    & 0.683    & 0.688   & 0.875    & {\textcolor{red} {0.958}} \\
				\multirow{-3}{*}{ViSal}  & MAE                      & 0.025   & 0.015 & {\textcolor{red} {0.010}} & 0.122 & 0.020 & 0.032 & 0.045 & 0.017                        & 0.026  & 0.021                        & 0.017 & 0.013                        & {\textcolor{red} {0.010}} & 0.038    & 0.107    & 0.105   & 0.035    & {\textcolor{red} {0.011}} \\
				
				\hline
				& $S_{measure}$            & 0.773   & 0.793 & 0.792                        & 0.712 & 0.742 & 0.818 & 0.715 & 0.791                        & {\textcolor{red} {0.873}}  & 0.786                        & 0.828 & 0.832                        & 0.846 & 0.682    & 0.615    & 0.619   & 0.765    & {\textcolor{red} {0.811}} \\
				& $F_{max}$                & 0.709   & 0.698 & 0.713                        & 0.690 & 0.670 & 0.742 & 0.669 & 0.734                        & {\textcolor{red} {0.833}}  & 0.704                        & 0.747 & 0.764                        & 0.791 & 0.648    & 0.506    & 0.482   & 0.702    & {\textcolor{red} {0.729}} \\
				\multirow{-3}{*}{VOS}    & MAE                      & 0.082   & 0.082 & 0.077                        & 0.162 & 0.099 & 0.078 & 0.097 & 0.075                        &{\textcolor{red} {0.051}}  & 0.091                        & 0.065 & 0.061                        & 0.060 & 0.106    & 0.162    & 0.172   & 0.089    & {\textcolor{red} {0.074}} \\
				\hline
				& $S_{measure}$            & 0.839   & 0.878 & 0.876                        & 0.794 & 0.857 & 0.851 & 0.809 & {\textcolor{red} {0.908}} & 0.872  & 0.879                        & 0.868 & 0.872                        & *                            & 0.747    & 0.651    & 0.659   & 0.803    & {\textcolor{red} {0.873}} \\
				& $F_{max}$                & 0.830   & 0.848 & 0.850                        & 0.797 & 0.816 & 0.821 & 0.767 & {\textcolor{red} {0.903}} & 0.859  & 0.865                        & 0.837 & 0.854                        & *                            & 0.727    & 0.571    & 0.564   & 0.792    & {\textcolor{red} {0.862}} \\
				\multirow{-3}{*}{FBMS}   & MAE                      & 0.060   & 0.044 & 0.040                        & 0.079 & 0.047 & 0.064 & 0.088 & {\textcolor{red} {0.027}} & 0.053  & 0.040                        & 0.040 & 0.038                        & *                            & 0.083    & 0.160    & 0.161   & 0.073    & {\textcolor{red} {0.042}} \\
				\hline

				& $S_{measure}$            & 0.854   & 0.829 & 0.859                        & 0.832 & 0.887 & 0.882 & 0.838 & 0.910                        & 0.886  & 0.892                        & 0.902 & 0.892                        & {\textcolor{red} {0.914}} & 0.795    & 0.688    & 0.676   & 0.846    & {\textcolor{red} {0.869}} \\
				& $F_{max}$                & 0.815   & 0.768 & 0.823                        & 0.783 & 0.861 & 0.855 & 0.783 & 0.892                        & 0.848  & 0.860                        & 0.880 & 0.865                        & {\textcolor{red} {0.900}} & 0.734    & 0.569    & 0.515   & 0.793    & {\textcolor{red} {0.844}} \\
				\multirow{-3}{*}{DAVIS}  & MAE                      & 0.038   & 0.057 & 0.045                        & 0.048 & 0.031 & 0.028 & 0.043 & 0.023                        & 0.027  & 0.028                        & 0.022 & 0.023                        & {\textcolor{red} {0.016}} & 0.044    & 0.100    & 0.103   & {\textcolor{red} {0.038}}    & 0.041\\
				\hline
				
		\end{tabular}}
	\end{center}
	
	\caption{Quantitative comparisons with state-of-the-art approaches on DAVSOD, ViSal, VOS, FBMS, and DAVIS across S-measure, max F-measure, and MAE. The best result are shown in red.}
	\label{table:state_of_the_art}
\end{table*}

\begin{figure*}[t]
	\centering
	\includegraphics[width=1\textwidth]{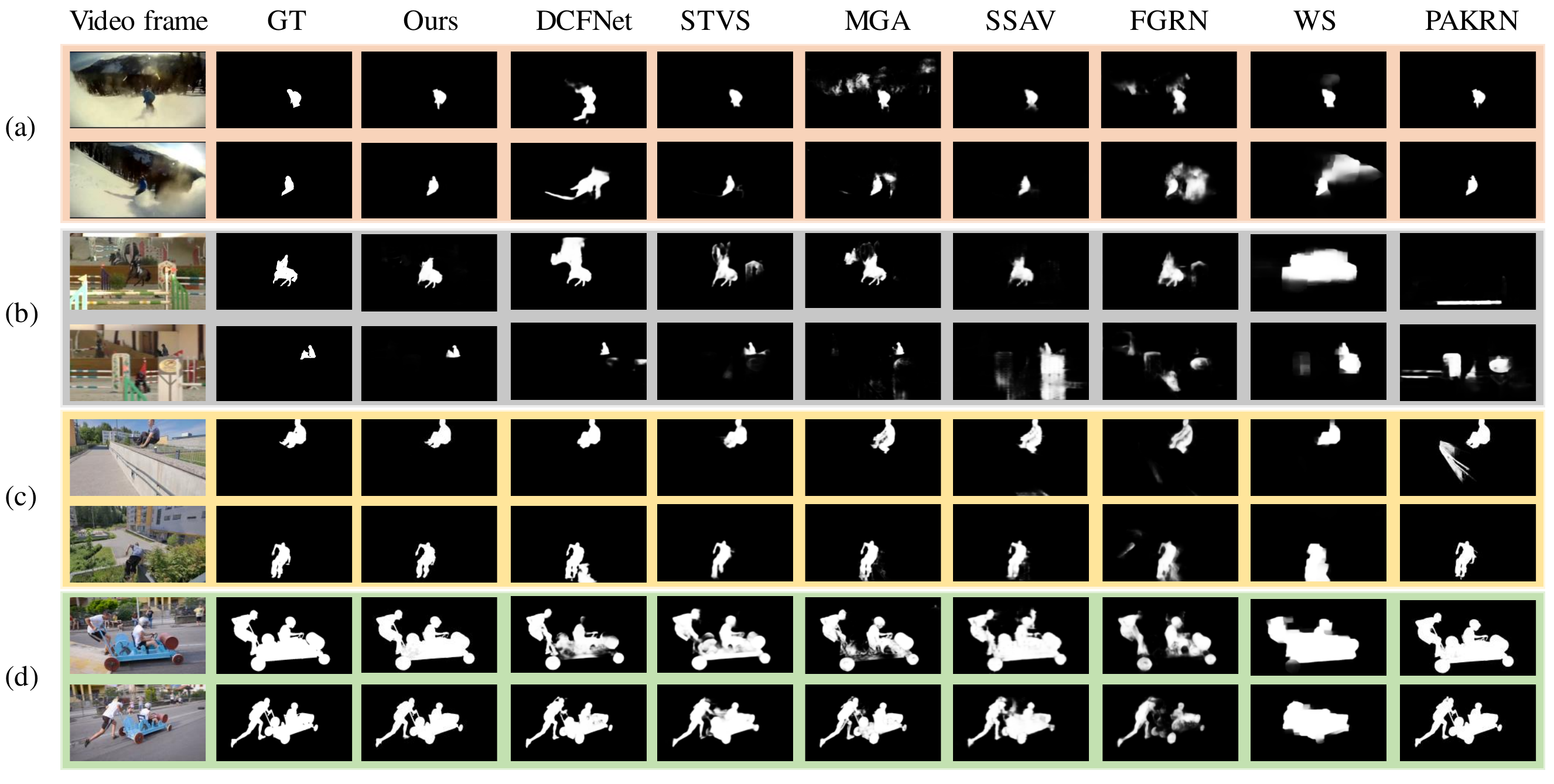} 
	\caption{Visual comparisons of our method with state-of-the-art methods. Each row shows the VSOD results of a frame. Each column indicates one algorithm. Our method consistently produces high-quality saliency maps and has more clear boundaries.}
	\label{fig:visualmap}
\end{figure*}

\subsection{Dataset}

We systematically test our proposed methods on five existing mainstream VSOD datasets, including ViSal~\cite{wang2015consistent}, VOS~\cite{li2017benchmark}, Freiburg-Berkeley motion segmentation dataset (FBMS)~\cite{ochs2013segmentation}, densely annotation video segmentation dataset (DAVIS)~\cite{perazzi2016benchmark}, and densely annotation VSOD dataset (DAVSOD)~\cite{fan2019shifting}. 

\noindent\textbf{-\textit{ViSal}} is the first dataset dedicated to the detection of video salient objects. It contains 17 video clips, each with a length of 30 to 100 frames, and 193 manually annotation frames. 

\noindent\textbf{-\textit{FBMS}} includes 59 video sequences and 720 labeled frames. 
\noindent\textbf{-\textit{VOS}}  consists of 200 videos. All objects and regions over 7,650 uniformly sampled keyframes are manually annotated and eye-tracking data of 23 subjects that free-view all videos are collected.

\noindent\textbf{-\textit{DAVIS}} has 50 video sequences and 3455 annotated frames. It contains multiple video object segmentation challenges, such as occlusion, motion blur, and appearance changes. Each video sequence is about 2-4 seconds. For every frame in the video, it provides pixel-level accuracy and manually creates the segmentation in a binary mask. 

\noindent\textbf{-\textit{DAVSOD}} is a larger dataset. It contains 226 video clips with the size of $640\times360$ for a total of 798 seconds and covers approximately 70 of the most common scenes/objects, with 23938 frames of object-level saliency annotations. It also provides an additional 40,000 frames of instance-level saliency annotations and short text descriptions. DAVSOD is randomly divided into 90 training sets, 46 validation sets, and 90 test sets. In addition, according to the difficulty of the task, the 90 test sets are further divided into 35 easy, 30 ordinary, and 25 difficult subsets.

\subsection{Evaluation Metrics}
Three indicators are used to evaluate our methods: mean absolute error (MAE)~\cite{perazzi2012saliency}, F-measure~\cite{achanta2009frequency}, Structural measurement (S-measure)~\cite{fan2017structure}. 


MAE is the mean value of absolute value of error between the ground truth and the saliency map. and it's defined as follows:
\begin{equation}
MAE=\frac1N\sum_{i=1}^N\vert(f_i-y_i)\vert,
\end{equation}
where $f_i$ denotes the predict saliency map value and $y_i$ denotes the groud truth value. 

F-measure is a weighted harmonic mean of precision and recall, and is defined as
\begin{equation}
F_\beta=\frac{\left(l+\beta^2\right)\;\cdot\;Precision\;\cdot\;Recall}{\beta^2\;\cdot\;Precision\;+\;Recall},
\end{equation}
We use the max f-measure over all thresholds from 0 to 255, denoted as $F_{max}$.

S-measure is used to evaluate the structural similarity between the saliency map and the ground truth map, which is defined as follows:
\begin{equation}
S\;=\;\alpha\;\ast\;S_0\;+\;(1-\alpha)\;\ast\;S_r,
\end{equation}
where $\alpha\in\lbrack0,1\rbrack$, and it is usually set to 0.5. $S_0$ means the object-aware structural similarity measure and $S_r$ means the region-aware structural similarity measure.

\subsection{Implementation Details}
Image salient object detection dataset DUTS-TR~\cite{duts_wang} with labels and VSOD datasets (\textit{e.g.} FBMS, DAVIS, and DAVSOD) without annotations are employed to implement our method. 
We evaluate VSOD methods on DAVIS, FBMS, ViSal, VOS, and DAVSOD benchmarks. 
We apply Adam optimizer~\cite{adam_king} with a weight decay of 1e-5 and learning rate of 5e-5 that is multiplied by 0.1 after 15 epochs to train two-stream locating network. 
We initialize the weights of the backbone of two-stream locating network from trained CLM and the model is traned for 24 epochs. All hyperparameter settings of the two-stream locating network are the same as CLM in PAKRN.
During training of two-stream locating network, for image data or first and last frames without adjacent frames, our optical flow branch inputs images with equal random values for each pixel. During testing, we also feed the optical flow branch of first and last frames without adjacent frames with images of fixed value.
The proposed framework is implemented based on PyTorch~\cite{pytorch_paszke}, a flexible, popular open source deep learning platform.
In our method, $T$ is set to 0.7.
All experiments are run with a mini-batch of 1 on an NVIDIA GeForce GTX Titan XP GPU.

\subsection{Comparison with the State-of-the-arts}
As shown in Table~\ref{table:state_of_the_art}, we compare our proposed method with existing three image salient object detection models (PoolNet~\cite{poolnet_liu}, EGNet~\cite{zhao2019egnet}, PAKRN~\cite{xu2021locate}), ten fully supervised VSOD models (SCOM~\cite{chen2018scom}, MBN~\cite{li2018unsupervised}, PDB~\cite{song2018pyramid}, FGRN~\cite{li2018flow},   MGA~\cite{li2019motion}, RCRN~\cite{yan2019semi}, SSAV~\cite{fan2019shifting}, PCSA~\cite{gu2020pyramid}, STVS~\cite{chen2021exploring}, and DCFN~\cite{zhang2021dynamic}), and four weakly or unsupervised models(SSOD~\cite{zhang2020weakly}, GF~\cite{wang2015consistent}, SAG~\cite{wang2015saliency}, WS~\cite{zhao2021weakly}) . For fair comparison, results of these methods are directly provided by authors or by their trained model and we use the same evaluation codes to test them. 
Our method does not have any post-processing.

For the image SOD methods, we can find that they perform well comparable to fully supervised models on ViSal and FBMS, especially in the ViSal dataset where the salient objects did not move dramatically and are easily distinguished in appearance. Our method also retains the advantages of image SOD models and shows comparable performance, which indicates that our method maintains good performance on the easy VSOD dataset. In several complicated datasets, especially on DAVSOD and VOS, image SOD methods don't perform well due to the inability to identify the evident motion of salient objects.
However, it can be seen that our method performs better, especially on the challenging dataset of DAVSOD, where our results are competitive to state-of-the-art fully supervised method. It demonstrates the effectiveness of our method in complex scenes, even without using pixel-wise annotations of videos. 
Further, compared with other weakly supervised and unsupervised models, our weakly supervised methods based on image data performs best on all datasets across all metrics. 
In addition, there is a semi-supervised video salient object detection method RCRN~\cite{yan2019semi}. Although it is called as a semi-supervised saliency detection method using pseudo-labels, it requires 20\% pixel-wise video annotations. The performance of our method is similar to that of RCRN, and Ours performs better on DAVSOD, ViSal, and FBMS, but worse on VOS ans DAVIS. 
Our method performs well and does not use any video annotation, which indicates the effectiveness and superiority of our method.
To further investigate the effectiveness of our proposed method, we show visual examples of our method with other approaches. As shown in Fig.~\ref{fig:visualmap}, our method can not accurately distinguish salient objects but also accomplish fine segmentation. Especially in the boundary, our method is able to segment small objects more precisely than other methods.


\begin{figure*}[t]
	\centering
	\includegraphics[width=1\textwidth]{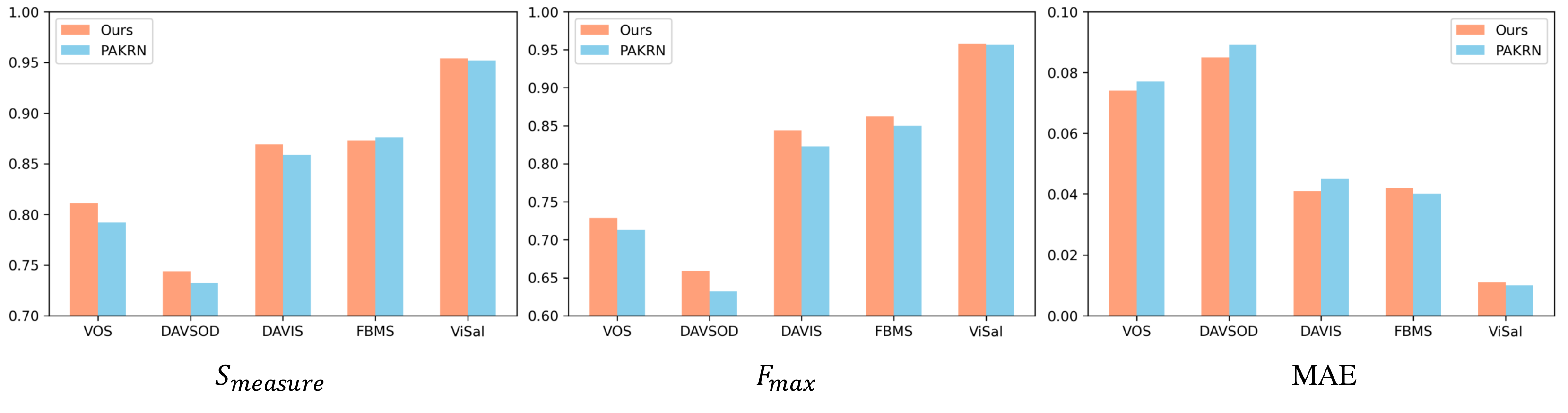}
	\caption{Comparison between our proposed method (red) and  PAKRN (blue) on VOS, DAVSOD, DAVIS, FBMS, and ViSal.}
	\label{fig:abalation_histo}
\end{figure*}

\begin{figure*}[t]
	\centering
	\includegraphics[width=1\textwidth]{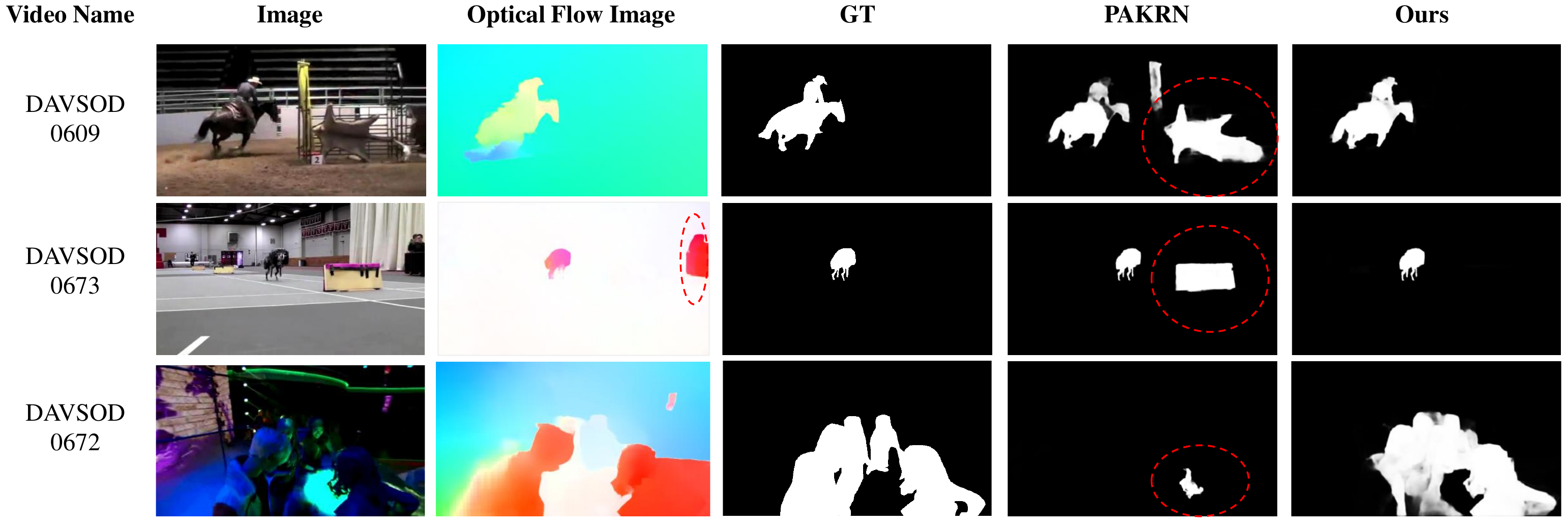} 
	\caption{Visual examples of our method trained by our extracted video frames with spatiotemporal location labels and PAKRN trained by image SOD dataset.}
	\label{fig:unsupervised_ablation}
\end{figure*}

\subsection{Ablation Study}
\subsubsection{Effectiveness of Strategies for Learning Video Knowledge from Image Data}
As Section~\ref{sec::method} described, our generated spatiotemporal location labels is based on CLM of PAKRN~\cite{xu2021locate}. To investigate the effectiveness of strategies for learning video knowledge from image data, we compare the result of ours (red) with PAKRN trained by image SOD dataset (shown in Fig.~\ref{fig:abalation_histo}).  The experiments are performed on DAVSOD, DAVIS, FBMS, and ViSal across three metrics ($S_m$, $F_{max}$, and MAE). 
It can be observed that our method performs significantly better on VOS, DAVSOD, DAVIS, but it is similar on ViSal and FBMS.
Because the data of ViSal and FBMS have evident static saliency. As shown in Table~\ref{table:state_of_the_art}, we can observe the superiority of PAKRN compared to the mainstream fully supervised VSDO approaches on ViSal and FBMS.
Such datasets can be well predicted by the model trained on the only image SOD dataset, so it is difficult for our method to show its effective role. 
In general, our method outperforms PAKRN, which can illustrate the effectiveness and reliability of our proposed VSOD method based on image SOD data.

To explore the reasons why our method works well, we show some examples in Fig.~\ref{fig:unsupervised_ablation}. 
In video 0673, the object in the red circle of optical flow image is dynamic saliency but not static saliency, and the object in the red circle of PAKRN is static saliency but not dynamic saliency, so they not the salient object in the video. In this case, our method can distinguish it well by fusing appearance information and motion information. 
In video 0672, the static appearance salient object is not the video salient object. Thus, PAKRN does not find the salient object at all, but our method can discover it by employing optical flow information.
In video 0609 with a complex background, it is difficult to distinguish salient objects by static saliency alone, but our method can effectively achieve it.
These outcomes explain the effectiveness of our method that focuses on both static appearance and motion information.

\begin{table}[]
	\begin{center}
		\setlength{\tabcolsep}{0.2mm}{
			\begin{tabular}{c|ccc|ccc}
				\hline
				DataSets     &           & DAVSOD        &       &           & FBMS          &       \\
				\hline
				Metric        & $S_{measure}$ & $F_{max}$ & MAE   & $S_{measure}$ & $F_{max}$ & MAE   \\
				\hline
				CLM w/o location labels     & 0.732     & 0.632         & 0.089 & 0.876     & 0.850         & 0.040 \\
				CLM w/ location labels      & 0.733     & 0.639         & 0.085 & 0.877     & 0.862         & 0.039 \\
				\hline
				\hline
				DataSets     &           & ViSal         &       &           & DAVIS         &       \\
				\hline
				Metric       & $S_{measure}$ & $F_{max}$ & MAE   & $S_{measure}$ & $F_{max}$ & MAE   \\
				\hline
				CLM w/o location labels     & 0.952     & 0.956         & 0.010 & 0.859     & 0.823         & 0.045 \\
				CLM w/ location labels      & 0.945     & 0.950         & 0.015 & 0.865     & 0.842         & 0.042 \\
				\hline
		\end{tabular}}
	\end{center}
	
	\caption{Comparison between coarse locating network (CLM) trained by our extracted video frames with spatiotemporal location labels and trained by image SOD dataset.}
	\label{table:AAAI_vs_clm}
\end{table}

\subsubsection{Effectiveness of generating spatiotemporal location labels}
As shown in Table~\ref{table:AAAI_vs_clm}, to evaluate the effectiveness of generating spatiotemporal location labels, we compare CLM  fine-tuned with and without our generated location labels. CLM w/ location labels refers to CLM fine-tuned by our extracted video frames with spatiotemporal location labels and CLM w/ location labels refers to CLM trained by only image SOD dataset. Both networks are the same, but they are fed with different training datasets. The former is fine-tuned by the selected frames and their spatiotemporal location maps after trained on the image SOD dataset, while the latter only uses the image SOD dataset.
We can observe that CLM with spatiotemporal location labels has significantly better results on DAVSOD, FBMS, and DAVIS across all metrics, which proves that our method can generated high-quality location labels. However, one noteworthy finding is that that CLM with spatiotemporal location labels performs not better on ViSal. The reason behind is that the location labels inevitably have a little bit of dirty data that can affect the model. For ViSal, a dataset where salient object detection can be easily achieved with a model trained with only image data, \textit{i.e.} static saliency objects are video saliency objects, the introduction of our spatiotemporal location labels slightly does more harm than good. However, the performance of CLM fine-tuned by spatiotemporal location labels on ViSal does not degrade much, which indicates that our generated location labels are still reliable.

\begin{table}[]
	\begin{center}
		\setlength{\tabcolsep}{0.3mm}{
			\begin{tabular}{c|ccc|ccc}
				\hline
				DataSets     &           & DAVSOD        &       &           & FBMS          &       \\
				\hline
				Metric        & $S_{measure}$ & $F_{max}$ & MAE   & $S_{measure}$ & $F_{max}$ & MAE   \\
				\hline
				CLM w/ location labels      & 0.733     & 0.639         & 0.085 & 0.877     & 0.862         & 0.039 \\
				Ours     & 0.744     & 0.659         & 0.085 & 0.873     & 0.862         & 0.042 \\
				\hline
				\hline
				DataSets     &           & ViSal         &       &           & DAVIS         &       \\
				\hline
				Metric       & $S_{measure}$ & $F_{max}$ & MAE   & $S_{measure}$ & $F_{max}$ & MAE   \\
				\hline
				CLM w/ location labels      & 0.945     & 0.950         & 0.015 & 0.865     & 0.842         & 0.042 \\
				Ours     & 0.954     & 0.958         & 0.011 & 0.869     & 0.844         & 0.041 \\
				\hline
		\end{tabular}}
	\end{center}
	
	\caption{Comparison between our proposed two-stream locating network and the coarse locating network (CLM) trained by our extracted video frames with spatiotemporal location labels.}
	\label{table:twostream_vs_clm}
\end{table}

\begin{table}[]
	\begin{center}
		\setlength{\tabcolsep}{0.8mm}{
			\begin{tabular}{c|ccc|ccc}
				\hline
				DataSets     &           & DAVSOD        &       &           & FBMS          &       \\
				\hline
				Metric       & $S_{measure}$ & $F_{max}$ & MAE   & $S_{measure}$ & $F_{max}$ & MAE   \\
				\hline
				w/o neighbor & 0.738     & 0.649         & 0.090 & 0.876     & 0.862         & 0.043 \\
				w/  neighbor   & 0.744     & 0.659         & 0.085 & 0.873     & 0.862         & 0.042 \\
				\hline
				\hline
				DataSets     &           & ViSal         &       &           & DAVIS         &       \\
				\hline
				Metric       & $S_{measure}$ & $F_{max}$ & MAE   & $S_{measure}$ & $F_{max}$ & MAE   \\
				\hline
				w/o neighbor & 0.951     & 0.954         & 0.013 & 0.865     & 0.834         & 0.044 \\
				w/ neighbor   & 0.954     & 0.958         & 0.011 & 0.869     & 0.844         & 0.041 \\
				\hline
		\end{tabular}}
	\end{center}
	
	\caption{Ablation study for tracking adjacent frames. W/o neighbor represents for the two-stream locating network fed with only high-saliency frames and no adjacent frames. W/o neighbor indicates that the two-stream locating network is trained by high-saliency frames and their adjacent frames.}
	\label{table:noneighber}
\end{table}

\begin{table}[t]
	\begin{center}
		\setlength{\tabcolsep}{0.7mm}{
			\begin{tabular}{c|ccc|ccc}
				\hline
				DataSets       &       & DAVSOD &       &       & FBMS  &       \\
				\hline
				Metric         & $S_{measure}$     & $F_{max}$      & MAE     & $S_{measure}$    & $F_{max}$     & MAE     \\
				\hline
				Optical flow (1) & 0.728 & 0.639  & 0.091 & 0.855 & 0.847 & 0.051 \\
				Optical flow (2) & 0.737 & 0.650  & 0.087 & 0.863 & 0.860 & 0.050 \\
				Optical flow (3) & 0.741 & 0.656  & 0.086 & 0.866 & 0.866 & 0.049 \\
				Optical flow (4) & 0.743 & 0.659  & 0.085 & 0.869 & 0.865 & 0.047 \\
				Optical flow (5) & 0.744 & 0.659  & 0.085 & 0.873 & 0.862 & 0.042 \\
				\hline
				\hline
				DataSets       &       & ViSal  &       &       & DAVIS &       \\
				\hline
				Metric         & $S_{measure}$     & $F_{max}$      & MAE     & $S_{measure}$     & $F_{max}$     & MAE     \\
				\hline
				Optical flow (1) & 0.944 & 0.949  & 0.017 & 0.873 & 0.838 & 0.037 \\
				Optical flow (2) & 0.949 & 0.954  & 0.014 & 0.879 & 0.843 & 0.036 \\
				Optical flow (3) & 0.949 & 0.953  & 0.015 & 0.879 & 0.840 & 0.038 \\
				Optical flow (4) & 0.953 & 0.956  & 0.012 & 0.879 & 0.841 & 0.039 \\
				Optical flow (5) & 0.954 & 0.958  & 0.011 & 0.869 & 0.844 & 0.041 \\
				\hline
		\end{tabular}}
	\end{center}
	
	\caption{Ablation study with different amounts of input optical flow images. Optical flow (n) represents for the n optical flow images are input in the dynamic branch.}
	\label{table:flow}
\end{table}

\subsubsection{Effectiveness of tracking adjacent frames}
To prove the effectiveness of the tracking adjacent frames strategy, we compare our two-stream locating network trained by high-saliency frames and their adjacent frame with by only high-saliency frames, as shown in Table~\ref{table:noneighber}. W/o neighbor represents for the two-stream locating network fed with only high-saliency frames and no adjacent frames. W/o neighbor indicates that the two-stream locating network is trained by high-saliency frames and their adjacent frames. 
Experiments shows that the two-stream locating network trained by spatiotemporal location labels including adjacent frames performs better on DAVSOD, ViSal, and DAVIS and not worse on FBMS. It demonstrates that tracking adjacent frames strategy is reliable and essential for better results.

\subsubsection{Effectiveness of the Proposed Two-stream locating Network}
To evaluate the effectiveness of the proposed two-stream locating network, we compare it with a single-stream network, \textit{i.e.} CLM, without dynamic optical flow branch, as shown in Table~\ref{table:twostream_vs_clm}. CLM w/ location labels refers to CLM trained by our extracted video frames with spatiotemporal location labels. For fair comparison, except for the network structure, other configurations are the same. As we can see, the two-stream locating network performs better than CLM on DAVSOD, ViSal, and DAVIS and slightly worse on FBMS. It demonstrates that our proposed two-stream locating network by fusing an dynamic optical flow branch is effective and necessary for better performance.

\subsubsection{Different amounts of input optical flow images}
As described in Section~\ref{sec::two-stream}, to alleviate the model's vulnerability to the quality of individual optical flow image, our proposed two-stream locating network employs five optical flow images as input. To validate the effectiveness of our solution, we compare our methods with different amounts of input optical flow images on DAVSOD, FBMS, ViSal, and DAVIS as shown in Table~\ref{table:flow}. We can see that when only one optical flow image is used as input, it performs significantly worse than others across most metrics. However, the performance of our method is improved when several optical flow images are employed as inputs, which proves that our multi-optical-flow input solution is effective in dealing with the vulnerability to the quality of individual optical flow image. In general, our method performs better when the number of input optical flow images is 4 and 5. Finally, we select five optical flow images as inputs that performs slightly better.

\section{Conclusion}
Facing the problem of time-consuming, inefficient, and costly acquisition of video saliency data by the human eye tracker and the few annotated video saliency datasets available, a novel SOD method via progressive framework is proposed to transfer the knowledge learned in the images to the VSOD task. The competitive results compared with the state-of-the-art fully supervised methods show that our approach is effective and successful. In addition, the methods of generating high-saliency location labels and tracking adjacent frames are simple. In future work, based on our solid framework, improving these two steps will enhance performance. Moreover, after obtaining high-quality spatiotemporal location labels, creating a network that can efficiently combine spatial and temporal information to achieve promising performance will be an important work. Overall, our method does not require video labels, so the more unlabeled video datasets collected for training and the richer the video scenes, our model will performs better.

\section*{Acknowledgment}
This work was partially supported by the National Key Research and Development Program of China (2020YFB1707700), the National Natural Science Foundation of China (62176235, 62036009, 61871350), and Zhejiang Provincial Natural Science Foundation of China (LY21F020026).

\ifCLASSOPTIONcaptionsoff
  \newpage
\fi

{
	\bibliographystyle{IEEEtran}
	\bibliography{egbib}

\begin{thebibliography}{10}
\providecommand{\url}[1]{#1}
\csname url@samestyle\endcsname
\providecommand{\newblock}{\relax}
\providecommand{\bibinfo}[2]{#2}
\providecommand{\BIBentrySTDinterwordspacing}{\spaceskip=0pt\relax}
\providecommand{\BIBentryALTinterwordstretchfactor}{4}
\providecommand{\BIBentryALTinterwordspacing}{\spaceskip=\fontdimen2\font plus
\BIBentryALTinterwordstretchfactor\fontdimen3\font minus
  \fontdimen4\font\relax}
\providecommand{\BIBforeignlanguage}[2]{{%
\expandafter\ifx\csname l@#1\endcsname\relax
\typeout{** WARNING: IEEEtran.bst: No hyphenation pattern has been}%
\typeout{** loaded for the language `#1'. Using the pattern for}%
\typeout{** the default language instead.}%
\else
\language=\csname l@#1\endcsname
\fi
#2}}
\providecommand{\BIBdecl}{\relax}
\BIBdecl

\bibitem{wang2015saliency}
W.~Wang, J.~Shen, and F.~Porikli, ``Saliency-aware geodesic video object
  segmentation,'' in \emph{Proceedings of the IEEE conference on computer
  vision and pattern recognition}, 2015, pp. 3395--3402.

\bibitem{rapantzikos2009spatiotemporal}
K.~Rapantzikos, N.~Tsapatsoulis, Y.~Avrithis, and S.~Kollias, ``Spatiotemporal
  saliency for video classification,'' \emph{Signal Processing: Image
  Communication}, vol.~24, no.~7, pp. 557--571, 2009.

\bibitem{evangelopoulos2009video}
G.~Evangelopoulos, A.~Zlatintsi, G.~Skoumas, K.~Rapantzikos, A.~Potamianos,
  P.~Maragos, and Y.~Avrithis, ``Video event detection and summarization using
  audio, visual and text saliency,'' in \emph{2009 IEEE International
  Conference on Acoustics, Speech and Signal Processing}.\hskip 1em plus 0.5em
  minus 0.4em\relax IEEE, 2009, pp. 3553--3556.

\bibitem{hong2015online}
S.~Hong, T.~You, S.~Kwak, and B.~Han, ``Online tracking by learning
  discriminative saliency map with convolutional neural network,'' in
  \emph{International conference on machine learning}, 2015, pp. 597--606.

\bibitem{hadizadeh2013saliency}
H.~Hadizadeh and I.~V. Baji{\'c}, ``Saliency-aware video compression,''
  \emph{IEEE Transactions on Image Processing}, vol.~23, no.~1, pp. 19--33,
  2013.

\bibitem{simon2009alerting}
L.~Simon, J.-P. Tarel, and R.~Br{\'e}mond, ``Alerting the drivers about road
  signs with poor visual saliency,'' in \emph{2009 IEEE Intelligent Vehicles
  Symposium}.\hskip 1em plus 0.5em minus 0.4em\relax IEEE, 2009, pp. 48--53.

\bibitem{wang2015}
S.~Wang, M.~Jiang, X.~M. Duchesne, E.~A. Laugeson, D.~P. Kennedy, R.~Adolphs,
  and Q.~Zhao, ``{Atypical Visual Saliency in Autism Spectrum Disorder
  Quantified through Model-Based Eye Tracking},'' \emph{Neuron}, vol.~88,
  no.~3, pp. 604--616, November 2015.

\bibitem{song2018pyramid}
H.~Song, W.~Wang, S.~Zhao, J.~Shen, and K.-M. Lam, ``Pyramid dilated deeper
  convlstm for video salient object detection,'' in \emph{Proceedings of the
  European conference on computer vision (ECCV)}, 2018, pp. 715--731.

\bibitem{fan2019shifting}
D.-P. Fan, W.~Wang, M.-M. Cheng, and J.~Shen, ``Shifting more attention to
  video salient object detection,'' in \emph{Proceedings of the IEEE conference
  on computer vision and pattern recognition}, 2019, pp. 8554--8564.

\bibitem{yan2019semi}
P.~Yan, G.~Li, Y.~Xie, Z.~Li, C.~Wang, T.~Chen, and L.~Lin, ``Semi-supervised
  video salient object detection using pseudo-labels,'' in \emph{Proceedings of
  the IEEE/CVF International Conference on Computer Vision}, 2019, pp.
  7284--7293.

\bibitem{li2019motion}
H.~Li, G.~Chen, G.~Li, and Y.~Yu, ``Motion guided attention for video salient
  object detection,'' in \emph{Proceedings of the IEEE International Conference
  on Computer Vision}, 2019, pp. 7274--7283.

\bibitem{ren2020tenet}
S.~Ren, C.~Han, X.~Yang, G.~Han, and S.~He, ``Tenet: Triple excitation network
  for video salient object detection,'' in \emph{European Conference on
  Computer Vision}.\hskip 1em plus 0.5em minus 0.4em\relax Springer, 2020, pp.
  212--228.

\bibitem{duts_wang}
L.~Wang, H.~Lu, Y.~Wang, M.~Feng, D.~Wang, B.~Yin, and X.~Ruan, ``Learning to
  detect salient objects with image-level supervision,'' in \emph{Proceedings
  of the IEEE Conference on Computer Vision and Pattern Recognition}, 2017, pp.
  136--145.

\bibitem{cheng2014global}
M.-M. Cheng, N.~J. Mitra, X.~Huang, P.~H. Torr, and S.-M. Hu, ``Global contrast
  based salient region detection,'' \emph{IEEE transactions on pattern analysis
  and machine intelligence}, vol.~37, no.~3, pp. 569--582, 2014.

\bibitem{perazzi2016benchmark}
F.~Perazzi, J.~Pont-Tuset, B.~McWilliams, L.~Van~Gool, M.~Gross, and
  A.~Sorkine-Hornung, ``A benchmark dataset and evaluation methodology for
  video object segmentation,'' in \emph{Proceedings of the IEEE Conference on
  Computer Vision and Pattern Recognition}, 2016, pp. 724--732.

\bibitem{rahtu2010segmenting}
E.~Rahtu, J.~Kannala, M.~Salo, and J.~Heikkil{\"a}, ``Segmenting salient
  objects from images and videos,'' in \emph{European conference on computer
  vision}.\hskip 1em plus 0.5em minus 0.4em\relax Springer, 2010, pp. 366--379.

\bibitem{liu2014superpixel}
Z.~Liu, X.~Zhang, S.~Luo, and O.~Le~Meur, ``Superpixel-based spatiotemporal
  saliency detection,'' \emph{IEEE transactions on circuits and systems for
  video technology}, vol.~24, no.~9, pp. 1522--1540, 2014.

\bibitem{xi2016salient}
T.~Xi, W.~Zhao, H.~Wang, and W.~Lin, ``Salient object detection with
  spatiotemporal background priors for video,'' \emph{IEEE Transactions on
  Image Processing}, vol.~26, no.~7, pp. 3425--3436, 2016.

\bibitem{chen2017video}
C.~Chen, S.~Li, Y.~Wang, H.~Qin, and A.~Hao, ``Video saliency detection via
  spatial-temporal fusion and low-rank coherency diffusion,'' \emph{IEEE
  Transactions on Image Processing}, vol.~26, no.~7, pp. 3156--3170, 2017.

\bibitem{zhao2021weakly}
W.~Zhao, J.~Zhang, L.~Li, N.~Barnes, N.~Liu, and J.~Han, ``Weakly supervised
  video salient object detection,'' in \emph{Proceedings of the IEEE/CVF
  Conference on Computer Vision and Pattern Recognition}, 2021, pp.
  16\,826--16\,835.

\bibitem{xu2021locate}
B.~Xu, H.~Liang, R.~Liang, and P.~Chen, ``Locate globally, segment locally: A
  progressive architecture with knowledge review network for salient object
  detection,'' in \emph{Proceedings of the AAAI Conference on Artificial
  Intelligence}, vol.~35, no.~4, 2021, pp. 3004--3012.

\bibitem{ochs2013segmentation}
P.~Ochs, J.~Malik, and T.~Brox, ``Segmentation of moving objects by long term
  video analysis,'' \emph{IEEE transactions on pattern analysis and machine
  intelligence}, vol.~36, no.~6, pp. 1187--1200, 2013.

\bibitem{wang2015consistent}
W.~Wang, J.~Shen, and L.~Shao, ``Consistent video saliency using local gradient
  flow optimization and global refinement,'' \emph{IEEE Transactions on Image
  Processing}, vol.~24, no.~11, pp. 4185--4196, 2015.

\bibitem{li2017benchmark}
J.~Li, C.~Xia, and X.~Chen, ``A benchmark dataset and saliency-guided stacked
  autoencoders for video-based salient object detection,'' \emph{IEEE
  Transactions on Image Processing}, vol.~27, no.~1, pp. 349--364, 2017.

\bibitem{borji2012exploiting}
A.~Borji and L.~Itti, ``Exploiting local and global patch rarities for saliency
  detection,'' in \emph{2012 IEEE conference on computer vision and pattern
  recognition}.\hskip 1em plus 0.5em minus 0.4em\relax IEEE, 2012, pp.
  478--485.

\bibitem{yan2013hierarchical}
Q.~Yan, L.~Xu, J.~Shi, and J.~Jia, ``Hierarchical saliency detection,'' in
  \emph{Proceedings of the IEEE conference on computer vision and pattern
  recognition}, 2013, pp. 1155--1162.

\bibitem{perazzi2012saliency}
F.~Perazzi, P.~Kr{\"a}henb{\"u}hl, Y.~Pritch, and A.~Hornung, ``Saliency
  filters: Contrast based filtering for salient region detection,'' in
  \emph{2012 IEEE conference on computer vision and pattern recognition}.\hskip
  1em plus 0.5em minus 0.4em\relax IEEE, 2012, pp. 733--740.

\bibitem{zhao2015saliency}
R.~Zhao, W.~Ouyang, H.~Li, and X.~Wang, ``Saliency detection by multi-context
  deep learning,'' in \emph{Proceedings of the IEEE conference on computer
  vision and pattern recognition}, 2015, pp. 1265--1274.

\bibitem{luo2017non}
Z.~Luo, A.~Mishra, A.~Achkar, J.~Eichel, S.~Li, and P.-M. Jodoin, ``Non-local
  deep features for salient object detection,'' in \emph{Proceedings of the
  IEEE Conference on computer vision and pattern recognition}, 2017, pp.
  6609--6617.

\bibitem{liu2018picanet}
N.~Liu, J.~Han, and M.-H. Yang, ``Picanet: Learning pixel-wise contextual
  attention for saliency detection,'' in \emph{Proceedings of the IEEE
  Conference on Computer Vision and Pattern Recognition}, 2018, pp. 3089--3098.

\bibitem{zhao2019pyramid}
T.~Zhao and X.~Wu, ``Pyramid feature attention network for saliency
  detection,'' in \emph{Proceedings of the IEEE Conference on Computer Vision
  and Pattern Recognition}, 2019, pp. 3085--3094.

\bibitem{hou2017deeply}
Q.~Hou, M.-M. Cheng, X.~Hu, A.~Borji, Z.~Tu, and P.~H. Torr, ``Deeply
  supervised salient object detection with short connections,'' in
  \emph{Proceedings of the IEEE Conference on Computer Vision and Pattern
  Recognition}, 2017, pp. 3203--3212.

\bibitem{zhang2017amulet}
P.~Zhang, D.~Wang, H.~Lu, H.~Wang, and X.~Ruan, ``Amulet: Aggregating
  multi-level convolutional features for salient object detection,'' in
  \emph{Proceedings of the IEEE International Conference on Computer Vision},
  2017, pp. 202--211.

\bibitem{hu2018recurrently}
X.~Hu, L.~Zhu, J.~Qin, C.-W. Fu, and P.-A. Heng, ``Recurrently aggregating deep
  features for salient object detection,'' in \emph{Thirty-second AAAI
  conference on artificial intelligence}, 2018.

\bibitem{zhang2018bi}
L.~Zhang, J.~Dai, H.~Lu, Y.~He, and G.~Wang, ``A bi-directional message passing
  model for salient object detection,'' in \emph{Proceedings of the IEEE
  Conference on Computer Vision and Pattern Recognition}, 2018, pp. 1741--1750.

\bibitem{zhang2018progressive}
X.~Zhang, T.~Wang, J.~Qi, H.~Lu, and G.~Wang, ``Progressive attention guided
  recurrent network for salient object detection,'' in \emph{Proceedings of the
  IEEE Conference on Computer Vision and Pattern Recognition}, 2018, pp.
  714--722.

\bibitem{li2018contour}
X.~Li, F.~Yang, H.~Cheng, W.~Liu, and D.~Shen, ``Contour knowledge transfer for
  salient object detection,'' in \emph{Proceedings of the European Conference
  on Computer Vision (ECCV)}, 2018, pp. 355--370.

\bibitem{chen2018reverse}
S.~Chen, X.~Tan, B.~Wang, and X.~Hu, ``Reverse attention for salient object
  detection,'' in \emph{Proceedings of the European Conference on Computer
  Vision (ECCV)}, 2018, pp. 234--250.

\bibitem{zhang2019salient}
P.~Zhang, W.~Liu, H.~Lu, and C.~Shen, ``Salient object detection with lossless
  feature reflection and weighted structural loss,'' \emph{IEEE Transactions on
  Image Processing}, vol.~28, no.~6, pp. 3048--3060, 2019.

\bibitem{wang2017video}
W.~Wang, J.~Shen, and L.~Shao, ``Video salient object detection via fully
  convolutional networks,'' \emph{IEEE Transactions on Image Processing},
  vol.~27, no.~1, pp. 38--49, 2017.

\bibitem{chen2018scom}
Y.~Chen, W.~Zou, Y.~Tang, X.~Li, C.~Xu, and N.~Komodakis, ``Scom:
  Spatiotemporal constrained optimization for salient object detection,''
  \emph{IEEE Transactions on Image Processing}, vol.~27, no.~7, pp. 3345--3357,
  2018.

\bibitem{li2018flow}
G.~Li, Y.~Xie, T.~Wei, K.~Wang, and L.~Lin, ``Flow guided recurrent neural
  encoder for video salient object detection,'' in \emph{Proceedings of the
  IEEE conference on computer vision and pattern recognition}, 2018, pp.
  3243--3252.

\bibitem{gu2020pyramid}
Y.~Gu, L.~Wang, Z.~Wang, Y.~Liu, M.-M. Cheng, and S.-P. Lu, ``Pyramid
  constrained self-attention network for fast video salient object detection,''
  in \emph{Proceedings of the AAAI Conference on Artificial Intelligence},
  vol.~34, no.~07, 2020, pp. 10\,869--10\,876.

\bibitem{chen2021exploring}
C.~Chen, G.~Wang, C.~Peng, Y.~Fang, D.~Zhang, and H.~Qin, ``Exploring rich and
  efficient spatial temporal interactions for real-time video salient object
  detection,'' \emph{IEEE Transactions on Image Processing}, vol.~30, pp.
  3995--4007, 2021.

\bibitem{zhang2021dynamic}
M.~Zhang, J.~Liu, Y.~Wang, Y.~Piao, S.~Yao, W.~Ji, J.~Li, H.~Lu, and Z.~Luo,
  ``Dynamic context-sensitive filtering network for video salient object
  detection,'' in \emph{Proceedings of the IEEE/CVF International Conference on
  Computer Vision}, 2021, pp. 1553--1563.

\bibitem{tang2018weakly}
Y.~Tang, W.~Zou, Z.~Jin, Y.~Chen, Y.~Hua, and X.~Li, ``Weakly supervised
  salient object detection with spatiotemporal cascade neural networks,''
  \emph{IEEE Transactions on Circuits and Systems for Video Technology},
  vol.~29, no.~7, pp. 1973--1984, 2018.

\bibitem{li2020plug}
Y.~Li, S.~Li, C.~Chen, A.~Hao, and H.~Qin, ``A plug-and-play scheme to adapt
  image saliency deep model for video data,'' \emph{IEEE Transactions on
  Circuits and Systems for Video Technology}, vol.~31, no.~6, pp. 2315--2327,
  2020.

\bibitem{fpn_lin}
T.-Y. Lin, P.~Doll{\'a}r, R.~Girshick, K.~He, B.~Hariharan, and S.~Belongie,
  ``Feature pyramid networks for object detection,'' in \emph{Proceedings of
  the IEEE conference on computer vision and pattern recognition}, 2017, pp.
  2117--2125.

\bibitem{zheng2019looking}
H.~Zheng, J.~Fu, Z.-J. Zha, and J.~Luo, ``Looking for the devil in the details:
  Learning trilinear attention sampling network for fine-grained image
  recognition,'' in \emph{Proceedings of the IEEE Conference on Computer Vision
  and Pattern Recognition}, 2019, pp. 5012--5021.

\bibitem{flow_gao}
C.~Gao, A.~Saraf, J.-B. Huang, and J.~Kopf, ``Flow-edge guided video
  completion,'' in \emph{European Conference on Computer Vision}.\hskip 1em
  plus 0.5em minus 0.4em\relax Springer, 2020, pp. 713--729.

\bibitem{render_butler}
D.~J. Butler, J.~Wulff, G.~B. Stanley, and M.~J. Black, ``A naturalistic open
  source movie for optical flow evaluation,'' in \emph{European conference on
  computer vision}.\hskip 1em plus 0.5em minus 0.4em\relax Springer, 2012, pp.
  611--625.

\bibitem{teed2020raft}
Z.~Teed and J.~Deng, ``Raft: Recurrent all-pairs field transforms for optical
  flow,'' \emph{arXiv preprint arXiv:2003.12039}, 2020.

\bibitem{2008Learning}
G.~Bradski and A.~Daebler, ``Learning opencv. computer vision with opencv
  library,'' \emph{University of Arizona Usa Since}, 2008.

\bibitem{poolnet_liu}
J.-J. Liu, Q.~Hou, M.-M. Cheng, J.~Feng, and J.~Jiang, ``A simple pooling-based
  design for real-time salient object detection,'' in \emph{Proceedings of the
  IEEE Conference on Computer Vision and Pattern Recognition}, 2019, pp.
  3917--3926.

\bibitem{47jia2020eml}
S.~Jia and N.~D. Bruce, ``Eml-net: An expandable multi-layer network for
  saliency prediction,'' \emph{Image and Vision Computing}, p. 103887, 2020.

\bibitem{48peters2005components}
R.~J. Peters, A.~Iyer, C.~Koch, and L.~Itti, ``Components of bottom-up gaze
  allocation in natural scenes,'' \emph{Journal of Vision}, vol.~5, no.~8, pp.
  692--692, 2005.

\bibitem{zhao2019egnet}
J.-X. Zhao, J.-J. Liu, D.-P. Fan, Y.~Cao, J.~Yang, and M.-M. Cheng, ``Egnet:
  Edge guidance network for salient object detection,'' in \emph{Proceedings of
  the IEEE/CVF International Conference on Computer Vision}, 2019, pp.
  8779--8788.

\bibitem{li2018unsupervised}
S.~Li, B.~Seybold, A.~Vorobyov, X.~Lei, and C.-C. Jay~Kuo, ``Unsupervised video
  object segmentation with motion-based bilateral networks,'' in
  \emph{Proceedings of the European Conference on Computer Vision (ECCV)},
  2018, pp. 207--223.

\bibitem{zhang2020weakly}
J.~Zhang, X.~Yu, A.~Li, P.~Song, B.~Liu, and Y.~Dai, ``Weakly-supervised
  salient object detection via scribble annotations,'' in \emph{Proceedings of
  the IEEE/CVF conference on computer vision and pattern recognition}, 2020,
  pp. 12\,546--12\,555.

\bibitem{achanta2009frequency}
R.~Achanta, S.~Hemami, F.~Estrada, and S.~Susstrunk, ``Frequency-tuned salient
  region detection,'' in \emph{2009 IEEE conference on computer vision and
  pattern recognition}.\hskip 1em plus 0.5em minus 0.4em\relax IEEE, 2009, pp.
  1597--1604.

\bibitem{fan2017structure}
D.-P. Fan, M.-M. Cheng, Y.~Liu, T.~Li, and A.~Borji, ``Structure-measure: A new
  way to evaluate foreground maps,'' in \emph{Proceedings of the IEEE
  international conference on computer vision}, 2017, pp. 4548--4557.

\bibitem{adam_king}
D.~P. Kingma and J.~Ba, ``Adam: A method for stochastic optimization,''
  \emph{arXiv preprint arXiv:1412.6980}, 2014.

\bibitem{pytorch_paszke}
A.~Paszke, S.~Gross, S.~Chintala, G.~Chanan, E.~Yang, Z.~DeVito, Z.~Lin,
  A.~Desmaison, L.~Antiga, and A.~Lerer, ``Automatic differentiation in
  pytorch,'' 2017.

\end{thebibliography}
}



\end{document}